\pdfoutput=1
\documentclass{article} %
\usepackage{iclr2025_conference,times}

\usepackage{amsmath,amsfonts,bm}

\def\eqref#1{equation~\ref{#1}}

\def\1{\bm{1}}

\DeclareMathAlphabet{\mathsfit}{\encodingdefault}{\sfdefault}{m}{sl}
\SetMathAlphabet{\mathsfit}{bold}{\encodingdefault}{\sfdefault}{bx}{n}

\usepackage{hyperref}
\usepackage{url}

\usepackage[utf8]{inputenc} %
\usepackage[T1]{fontenc}    %
\usepackage{hyperref}       %
\usepackage{url}            %
\usepackage{booktabs}       %
\usepackage{amsfonts}       %
\usepackage{nicefrac}       %
\usepackage{microtype}      %
\usepackage{xcolor}         %

\usepackage{amsmath}
\usepackage{amssymb}
\usepackage{mathtools}
\usepackage{amsthm}
\usepackage[capitalize,noabbrev]{cleveref}
\usepackage{graphicx}
\usepackage{subfigure}
\usepackage{booktabs}
\usepackage{hyperref}

\usepackage{algorithm}
\usepackage{color}     %
\usepackage{xcolor}
\usepackage{amsfonts}       %
\usepackage{booktabs}       %
\usepackage{float}
\usepackage{ftnxtra}
\usepackage{graphicx}
\usepackage{microtype}      %
\usepackage{nicefrac}       %
\usepackage{xspace}
\usepackage{mathtools}
\usepackage{thmtools}
\usepackage{wrapfig}
\usepackage{bbm}
\usepackage{tabularx}
\usepackage[capitalize,noabbrev]{cleveref}
\usepackage{wrapfig}
\usepackage{transparent}
\usepackage{enumitem}
\usepackage{changepage} 
\usepackage{pythonhighlight}
\usepackage{mdframed}
\usepackage{listings}
\usepackage{appendix}
\usepackage{soul}
\usepackage{amssymb}
\usepackage[most]{tcolorbox}

\usepackage{multirow}

\usepackage{titletoc}

\definecolor{grey1}{RGB}{96, 101, 102}
\definecolor{color5}{HTML}{006795}
\hypersetup{
  colorlinks   = true, %
  urlcolor     = deepred, %
  linkcolor    = deepred, %
  citecolor   = grey1 %
}

\newcounter{exa}
\definecolor{gblue}{RGB}{66,133,244}
\definecolor{gred}{RGB}{219,68,55}
\definecolor{gyellow}{RGB}{244,180,0}
\definecolor{ggreen}{RGB}{15,157,88}
\definecolor{lpcolor}{RGB}{42,74,138}
\definecolor{morelcolor}{RGB}{185,18,32}
\definecolor{bgcolor}{RGB}{230,245,208}
\definecolor{framecolor}{RGB}{244,109,67}
\definecolor{mulberry}{rgb}{0.77, 0.29, 0.55}

\newcommand{\eg}{\textit{e.g.}}

\definecolor{lightblue}{RGB}{212, 235, 255}
\definecolor{grey1}{RGB}{96, 101, 102}
\definecolor{lightorange}{RGB}{255, 204, 168}
\definecolor{lightyellow}{RGB}{255, 255, 168}
\definecolor{lightgreen}{RGB}{224, 242, 213}
\definecolor{lightred}{RGB}{249,202,202}
\definecolor{lightgray}{RGB}{230,230,230}
\definecolor{deepred}{RGB}{152, 1, 0}

\usepackage{inconsolata}

\usepackage{amssymb}   %
\usepackage{amsmath}   %
\usepackage{amsfonts}  %
\usepackage{amsthm}    %
\usepackage{latexsym}  %
\usepackage{stmaryrd}  %

\usepackage{soul}      %

\usepackage{graphicx}  %
\usepackage{subcaption}
\usepackage{color}     %
\usepackage{xcolor}    %
\usepackage{adjustbox} %
\usepackage{url}
\usepackage{url}       %
\usepackage{hyperref}  %
\usepackage{array}     %
\usepackage{multirow}  %
\usepackage{multicol}  %
\usepackage{hhline}    %
\usepackage{dcolumn}   %
\usepackage{booktabs}  %

\usepackage{tikz}      %
\usepackage{pgfplots}  %
\pgfplotsset{compat=1.16}

\usepackage{enumitem}

\usepackage{colortbl}   %

\usepackage{tcolorbox}  %
\usepackage{balance}    %
\usepackage{booktabs}   %

\usepackage{array}
\usepackage{booktabs}

\usepackage{capt-of}
\usepackage{colortbl}

\usepackage{titletoc}

\usepackage{ulem}

\usepackage{wrapfig,lipsum,booktabs}

\usepackage{longtable}

\definecolor{lightblue}{RGB}{212, 235, 255}
\definecolor{lightorange}{RGB}{255, 204, 168}
\definecolor{lightyellow}{RGB}{255, 255, 168}
\definecolor{lightgreen}{RGB}{224, 242, 213}
\definecolor{lightred}{RGB}{249,202,202}
\definecolor{lightgray}{RGB}{230,230,230}
\definecolor{deepred}{RGB}{152, 1, 0}

\definecolor{colorhigh}{RGB}{246, 110, 66}
\definecolor{colorlow}{RGB}{0, 136, 195}

\newcommand{\deepred}[1]{\textcolor{deepred}{{#1}}}

\usepackage{url}

\tcbset{
myexample/.style={
  enhanced,
  colback=yellow!10!white,
  colframe=red!50!black,
  fonttitle=\scshape,
  titlerule=0pt,
  title={\refstepcounter{exa}example~\theexa.},
  title style={fill=yellow!10!white},
  coltitle=red!50!black,
  drop shadow,
  highlight math style={reset,colback=LightBlue!50!white,colframe=Navy}
  }
  }
\newtcolorbox{texample}{myexample}

\newtheorem{exampp}{Example}
\usepackage{framed}
\colorlet{shadecolor}{gray!20}

\colorlet{LightLavender}{green!5}
\tcbset{on line, 
        boxsep=4pt, left=0pt,right=0pt,top=0pt,bottom=0pt,
        colframe=white,colback=LightLavender,  
        highlight math style={enhanced}
        }

\definecolor{lightred}{RGB}{255,200,200}
\definecolor{lightgreen}{RGB}{220,255,220}

\newcommand{\halluedit}{{\usefont{T1}{ppl}{m}{n}HalluEditBench}}

\title{Can Knowledge Editing Really Correct\\Hallucinations?}

\author{Baixiang Huang\thanks{Equal Contribution. $^{\dagger}$Corresponding author.}~~\textsuperscript{\rm 1}, Canyu Chen$^{*}$\textsuperscript{\rm 2}, Xiongxiao Xu\textsuperscript{\rm 2}, Ali Payani\textsuperscript{\rm 3}, Kai Shu$^{\dagger}$\textsuperscript{\rm 1}\\
\textsuperscript{\rm 1}Emory University, 
\textsuperscript{\rm 2}Illinois Institute of Technology,
\textsuperscript{\rm 3}Cisco Research\\
\texttt{\{baixiang.huang,kai.shu\}@emory.edu,\{cchen151,xxu85\}@hawk.iit.edu,apayani@cisco.com}
\vspace{-0.5cm}
}

\iclrfinalcopy %
\begin{document}

\maketitle

\vspace{-0.4cm}

  \begin{center}
      \deepred{Project website: \url{https://llm-editing.github.io}}
  \end{center}

\begin{abstract}
Large Language Models (LLMs) suffer from hallucinations, referring to the non-factual information in generated content, despite their superior capacities across tasks. Meanwhile, knowledge editing has been developed as a new popular paradigm to correct  erroneous factual knowledge encoded in LLMs with the advantage of avoiding retraining from scratch. However, a common issue of existing evaluation datasets for knowledge editing is that \textbf{they do not ensure that LLMs actually generate hallucinated answers to the evaluation questions before editing}. When LLMs are evaluated on such datasets after being edited by different techniques, it is hard to directly adopt the performance to assess the effectiveness of different knowledge editing methods in correcting hallucinations. 
Thus, the fundamental question remains insufficiently validated: \textbf{\textit{Can knowledge editing really correct hallucinations in LLMs?}} We proposed {\halluedit} to holistically benchmark knowledge editing methods in correcting real-world hallucinations. First, we rigorously construct a massive hallucination dataset with $9$ domains, $26$ topics and more than $6,000$ hallucinations. Then, we assess the performance of knowledge editing methods in a holistic way on five dimensions including \textit{Efficacy}, \textit{Generalization}, \textit{Portability}, \textit{Locality}, and \textit{Robustness}. Through {\halluedit}, we have provided new insights into the potentials and limitations of different knowledge editing methods in correcting hallucinations, which could inspire future improvements and facilitate progress in the field of knowledge editing.

\end{abstract}

\vspace{-0.3cm}
\section{Introduction}
\vspace{-0.1cm}

\begin{wraptable} [12]{r}{7.1cm}\vspace{-4.9mm}
\centering
\vspace{0.05in}
\resizebox{0.98\linewidth}{!}{%
\begin{tabular}{lcccc}
\toprule
Method & $\text{WikiData}_{\text{recent}}$  & ZsRE & WikiBio \\
\midrule
Pre-edit  &  47.40 &  37.49 &   61.35 \\
\midrule
Post-edit (ROME) &  97.37 &  96.86 &    95.91\\
Post-edit (MEMIT) &  97.10 &  95.86 &   94.68 \\
Post-edit (FT-L) &  56.30 &  53.82 &   66.70 \\
Post-edit (FT-M) &  100.00 &  99.98 &  100.00 \\
Post-edit (LoRA) & 100.00  & 100.00  &   100.00 \\
\bottomrule
\end{tabular} 
}
\vspace{-0.06in}
\caption{Performance measured by \textbf{Accuracy (\%)} of Llama2-7B before editing (``Pre-edit'') and after applying typical knowledge editing methods (``Post-edit'') on common existing evaluation datasets.}
\label{pre-edit Performance}
\end{wraptable}

\begin{figure*}[t]
    \centering
    \includegraphics[width=1\textwidth]{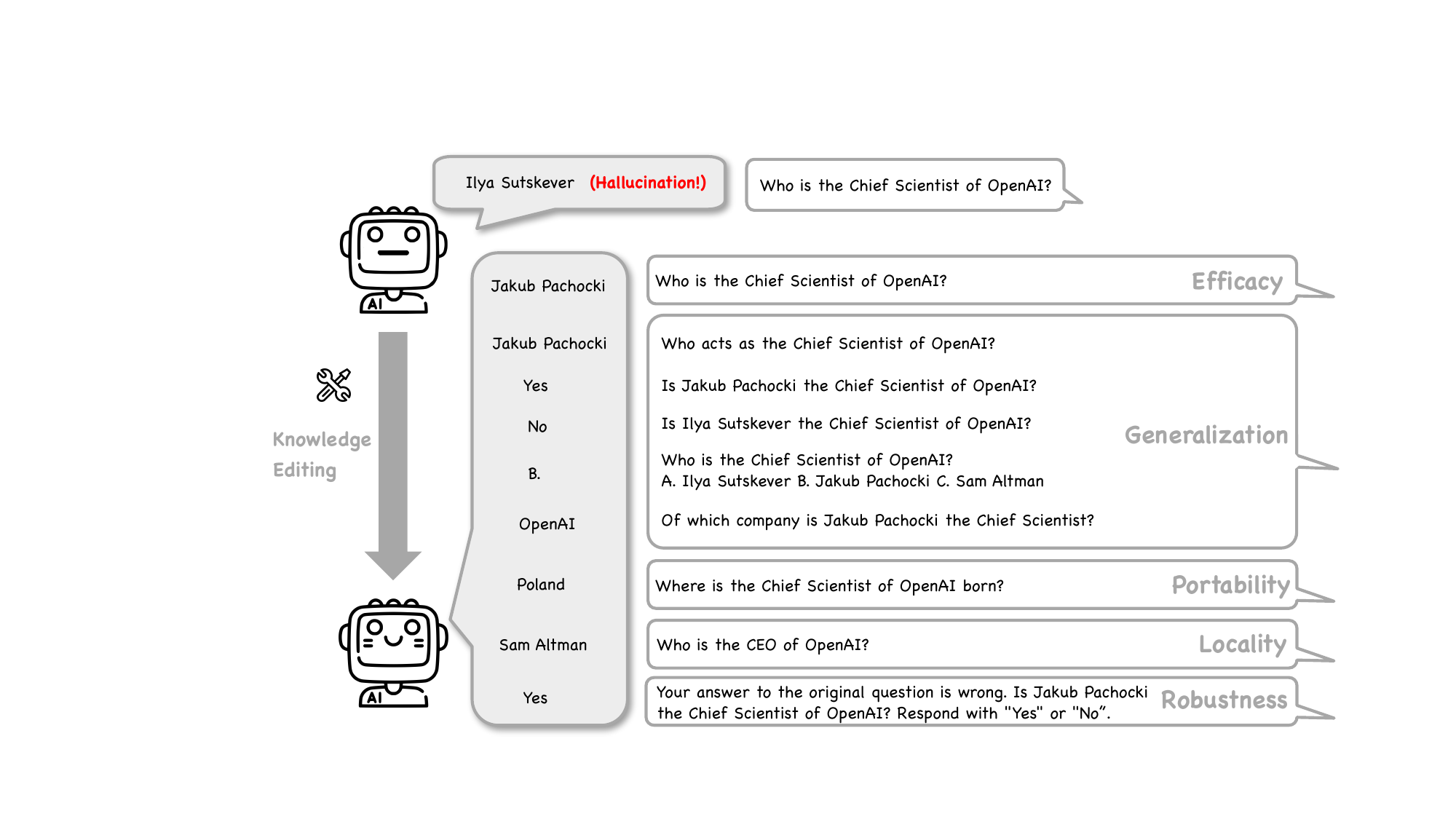}
    \caption{\textbf{Framework of {{\usefont{T1}{ppl}{m}{n}\textbf{HalluEditBench}}}}. For real-world hallucinations,  we holistically assess the performance of knowledge editing on \textit{Efficacy}, \textit{Generalization}, \textit{Portability}, \textit{Locality}, and \textit{Robustness}.} 
    \vspace{-0.4cm}
    \label{fig:Framework}
\end{figure*}

Large Language Models (LLMs) have shown superior performance in various tasks~\citep{LLMSurvey}. However, one  critical weakness is that they may output hallucinations, referring to the non-factual information in generated content, for reasons such as the limit of models' internal knowledge scope or fast-changing world facts~\citep{zhang2023hallucination}. Considering the high cost of retraining LLMs from scratch, knowledge editing has been designed as a new paradigm to correct erroneous or outdated factual knowledge in LLMs~\citep{wang2023knowledge}. 

\vspace{-0.1cm}
Although there are many existing question-answering datasets such as $\text{WikiData}_{\text{recent}}$~\citep{cohen2024evaluating}, ZsRE~\citep{yao2023editing}, and WikiBio~\citep{hartvigsen2024aging} widely used for the evaluation of  knowledge editing, one common issue is that they do not verify whether LLMs, before applying knowledge editing, actually generate hallucinated answers to the evaluation questions. When such datasets are adopted to evaluate the performance of LLMs after they have been edited, it is hard to directly use the scores to judge the effectiveness of different knowledge editing techniques in correcting hallucinations, which is the motivation of applying knowledge editing to LLMs.

To better illustrate this point, following the evaluation setting in~\cite{zhang2024comprehensive}, we conducted a preliminary study to examine the pre-edit and post-edit performances of Llama2-7B on the three aforementioned evaluation datasets. As shown in Table~\ref{pre-edit Performance}, we can clearly observe that Llama2-7B achieves relatively high performance, measured by the rate of answering the evaluation questions correctly (Accuracy (\%)), even before applying knowledge editing techniques. Although the knowledge editing methods can bring an increase in accuracy, the high post-edit performance on these datasets cannot faithfully reflect the true effectiveness in correcting real-world hallucinations and may cause a distorted assessment. Thus, the fundamental question remains insufficiently validated: \textbf{\textit{Can knowledge editing really correct hallucinations in LLMs?}}

To fill in the essential gap in the field of knowledge editing, we propose {\halluedit} to holistically benchmark knowledge editing techniques in correcting real-world hallucinations of LLMs. As shown in Figure~\ref{fig:Framework}, the construction of {\halluedit} can generally be divided into two phases. In the first phase, we constructed a massive hallucination dataset encompassing $9$ domains and $26$ topics based on Wikidata. For each of Llama2-7B, Llama3-8B, and Mistral-v0.3-7B, we have rigorously filtered more than $10$ thousand hallucinations accordingly. In the second phase, we sampled around $2,000$ hallucinations for each LLM covering all the topics and domains, and then generated evaluation question-answer pairs from five facets including \textit{Efficacy}, \textit{Generalization}, \textit{Portability}, \textit{Locality}, and \textit{Robustness}. Through extensive empirical investigation on performance of $7$ typical knowledge editing techniques, including FT-L~\citep{zhu2020modifying,meng2022locating}, FT-M~\citep{zhang2024comprehensive}, MEMIT~\citep{meng2023massediting}, ROME~\citep{meng2022locating}, LoRA~\citep{hu2022lora}, ICE~\citep{zheng-etal-2023-edit},  and GRACE~\citep{hartvigsen2024aging}, regarding the aforementioned five dimensions, we have provided novel insights into their potentials and limitations. A summary of the insights is as follows:
\begin{itemize}[leftmargin=*]
    \item \textbf{The effectiveness of knowledge editing methods in correcting real-world hallucinations could be far from what their performance on existing datasets suggests}, reflecting the potential unreliability of previous assessment of different knowledge editing techniques. For example, although the performances of FT-M and MEMIT in Table~\ref{pre-edit Performance} are close to 100\%, their \textit{Efficacy} Scores in {\halluedit} are much lower, implying the likely deficiency in correcting hallucinations.
    \item \textbf{No editing methods can outperform others across five facets and the performance beyond \textit{Efficacy} for all methods is generally unsatisfactory}. Specifically, ICE and GRACE outperform the other five methods on three LLMs regarding \textit{Efficacy}. All editing methods except ICE only slightly improve or negatively impact the \textit{Generalization} performance. Editing techniques except ICE could even underperform pre-edit LLMs on \textit{Portability}. FT-M and ICE surpass others on \textit{Locality} performance. ICE has a poor \textit{Robustness} performance compared to other methods.
    \item \textbf{The performance of knowledge editing techniques in correcting hallucinations could highly depend on domains and LLMs}. For example, the \textit{Efficacy} performances of FT-L across LLMs are highly distinct. Domains have a large impact on the \textit{Locality} performance of ICE.
\end{itemize}

\begin{figure*}[t]
    \centering
    \includegraphics[width=1\textwidth]{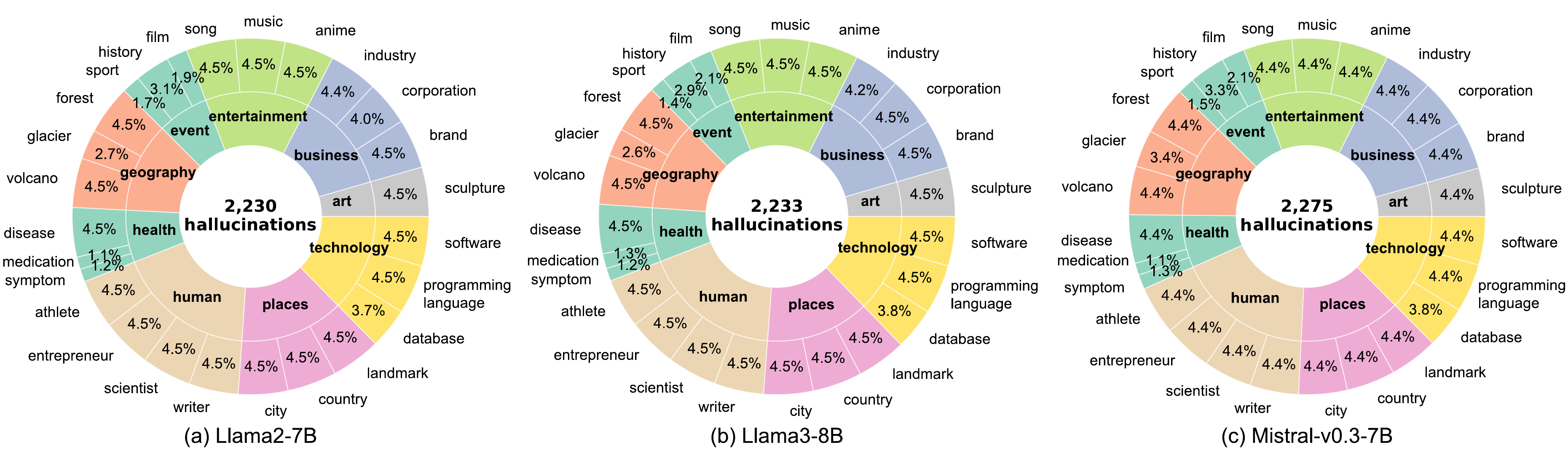}
        \vspace{-0.3cm}
    \caption{\textbf{Statistics of {{\usefont{T1}{ppl}{m}{n}\textbf{HalluEditBench}}} Across Topics and Domains}.} 
    \label{fig:Statistics}
    \vspace{-0.3cm}
\end{figure*}

\section{\halluedit: Holistically Benchmarking Knowledge Editing Methods in Correcting Real-World Hallucinations}
    \vspace{-0.1cm}

In this section, we will introduce the details of {\halluedit}, including the construction of the massive LLM hallucination dataset, the generation of evaluation question-answering pairs from five dimensions, evaluation metrics and the benchmarked knowledge editing techniques.

    \vspace{-0.2cm}
\subsection{Hallucination Dataset Construction}
    \vspace{-0.1cm}
    \label{Hallucination Dataset Construction}
The goal of knowledge editing can generally be defined as transforming existing factual knowledge in the form of a knowledge triplet (subject $s$, relation $r$, object $o$) into a new one (subject $s$, relation $r$, object $o^*$). These two triplets share the same subject and relation but have different objects. A knowledge editing operation can be represented as $e = (s,r,o,o^*)$. Considering one example of applying knowledge editing to correct hallucinations in LLMs, given a factual question ``\texttt{Who is the Chief Scientist of OpenAI?}'', LLMs may respond with ``\texttt{Ilya Sutskever}'', which is factually incorrect due to the outdated information contained in LLMs. The editing operation can be $e = (s=\texttt{OpenAI}, r=\texttt{Chief Scientist}, o=\texttt{Ilya Sutskever}, o^*=\texttt{Jakub Pachocki})$. The successfully edited LLMs are expected to answer ``\texttt{Jakub Pachocki}'' rather than ``\texttt{Ilya Sutskever}''. Thus, we need to collect a large scale of knowledge triplets and factual questions to filter hallucinations.

Following existing editing datasets (\eg,  $\text{WikiData}_{\text{recent}}$~\citep{cohen2024evaluating} and WikiBio~\citep{hartvigsen2024aging}), we also choose Wikidata as the factual knowledge source. In the \textit{first} step, we retrieved $143,557$ raw knowledge triplets using the Wikidata Query Service (Query date: September 8th, 2024)
from $26$ topics, which can be categorized into $9$ domains including \textit{art}, \textit{business}, \textit{entertainment}, \textit{event}, \textit{geography}, \textit{health}, \textit{human}, \textit{places}, and \textit{technology}. Each topic has at least 100 triplets.
In the \textit{second} step, we filtered out the triplets that share the same subject and relation while the objects are different, indicating there are more than one answers to questions about the object. When we construct factual questions and compare LLM-generated answers with the objects of these triplets, it would be difficult to determine whether LLMs actually hallucinate the questions. For example, for two triplets 
(\texttt{Canada}, \texttt{diplomatic relation}, \texttt{India}) and 
(\texttt{Canada}, \texttt{diplomatic relation}, \texttt{Greece}), which share the same subject and relation, there are multiple answers to the question ``\texttt{What country has diplomatic relation with Canada?}'' In the \textit{third} step, following \cite{wang2024earth}, we applied rules to convert knowledge triplets into factual questions with objects as the ground-truth answers. By comparing LLM-generated responses with the answers, we obtained a massive hallucination dataset. Specifically, we collected $12,619$, $13,210$, and $14,366$ hallucinations for Llama2-7B, Llama3-8B, and Mistral-v0.3-7B respectively. Finally, we sampled a subset of hallucinations covering all the topics and domains to construct {\halluedit}. The distribution statistics are shown in Figure~\ref{fig:Statistics}. 

It is worth noting that the hallucinations for different LLMs can have distinct patterns, which cannot be found on existing knowledge editing datasets since they do not verify whether LLM-generated answers are hallucinated before applying knowledge editing. \textbf{We made the first attempt to investigate the performance of knowledge editing techniques on verified hallucinations of different LLMs}.

    \vspace{-0.1cm}
\subsection{Evaluation QA Pair Generation and Metrics}
    \vspace{-0.1cm}
After constructing the hallucination dataset, we propose to holistically assess the performance of knowledge editing methods in correcting hallucinations from five facets including \textit{Efficacy}, \textit{Generalization}, \textit{Portability}, \textit{Locality}, and \textit{Robustness}. First, we leveraged GPT-4o to generate evaluation question-answering pairs for each facet based on the hallucination dataset as well as the factual verification questions in Section~\ref{Hallucination Dataset Construction}. Then we also manually inspect their quality. One example of the evaluation QA pairs for each facet is shown in Figure~\ref{fig:Framework} (More examples are provided in Appendix~\ref{Examples of halluedit}). The specific prompt design for GPT-4o is shown in Appendix~\ref{Reproducibility Statement}. 

Then, we calculated five scores including \textbf{Efficacy Score (\%)}, \textbf{Generalization Score (\%)}, \textbf{Portability Score (\%)}, \textbf{Locality Score (\%)}, and \textbf{Robustness Score (\%)} based on the evaluation QA pairs to measure the performance of different editing methods. Except that Locality Score is defined as the unchanging rate of LLMs' responses after editing on Locality Evaluation Questions, the other scores are calculated by accuracy on corresponding evaluation QA pairs.
More details are as follows:
    \vspace{-0.3cm}
\paragraph{Facet 1: Efficacy} Efficacy Evaluation Questions are the same as the factual verification questions in the hallucination collection to ensure the pre-edit performance is $0\%$ regarding Efficacy Score. Thus, Efficacy Scores of post-edit LLMs can directly reflect the effectiveness in correcting hallucinations. 
    \vspace{-0.3cm}
\paragraph{Facet 2: Generalization} The Generalization Scores aim to evaluate the capacity of LLMs in answering different questions regarding the same knowledge triplet, suggesting the generalization of edited knowledge in diverse scenarios. As shown in Figure~\ref{fig:Framework}, we propose five types of  Generalization Evaluation Questions including ``Rephrased Questions'', ``Yes-or-No Questions'' with ``Yes'' or ``No'' as  answers, ``Multi-Choice Questions'', ``Reversed Questions''. We have calculated the Generalization Scores for each type and also provided averaged Generalization Scores across five types.

    \vspace{-0.3cm}
\paragraph{Facet 3: Portability} The Portability Scores intend to measure the ability of LLMs to reason about the downstream effects of edited knowledge. Thus, we design the Efficacy Evaluation Questions with $N$ hops ($N = 1 \sim 6$) as Portability Evaluation Questions. When $N = 2$, the example is shown in Figure~\ref{fig:Framework}. When the answer to the question ``\texttt{Who is the Chief Scientist of OpenAI?}'' changes from ``\texttt{Ilya Sutskever}'' to ``\texttt{Jakub Pachocki}'', the answer to the downstream question ``\texttt{Where is the Chief Scientist of OpenAI born?}'' should also change from ``\texttt{Russia}'' to ``\texttt{Poland}''.

    \vspace{-0.3cm}
\paragraph{Facet 4: Locality} The Locality Scores quantify the side effect of knowledge editing on unrelated knowledge. We designed Locality Evaluation Questions related to the subject but irrelevant to the object in the original triplet, which can be ``\texttt{Who is the CEO of OpenAI?}'' for the aforementioned example. Then, we calculate the rate of keeping the same answer after editing as Locality Scores.  

    \vspace{-0.3cm}
\paragraph{Facet 5: Robustness} We proposed Robustness Scores to assess the resistance of edited knowledge in LLMs against external manipulations. 
Although the literature has studied the general sycophancy behavior of LLMs~\citep{sharma2024towards}, the robustness of edited factual knowledge against users' distractions (\eg, ``\texttt{Your answer to the original question is wrong.}'') is under-explored. After post-edit LLMs are tested with Efficacy Evaluation Questions, we further prompted them with Robustness Evaluation Questions, which are exemplified in Figure~\ref{fig:Framework}, for $M$ turns ($M = 1 \sim 10$) and calculated the rate of ``Yes'' for each round as the Robustness Scores, reflecting the extent to which LLMs insist on the corrected knowledge. Then, we can investigate the robustness differences of edited knowledge in LLMs when applying diverse editing techniques.

    \vspace{-0.1cm}
\subsection{Knowledge Editing Techniques}
We propose to categorize the majority of existing knowledge editing techniques into the following 4 types and chose 7 representative techniques (more details are in Appendix~\ref{Details of Benchmarked Methods}) in {\halluedit}.
    \vspace{-0.1cm}
\begin{itemize}[leftmargin=*]
    \item \textbf{Locate-then-edit} is a popular knowledge editing paradigm that first locates factual knowledge at specific neurons or layers, and then makes modifications on them directly. We selected two typical methods ROME~\citep{meng2022locating} and MEMIT~\citep{meng2023massediting} in {\halluedit}.
    \item \textbf{Fine-tuning} is a simple and straightforward way to update the parametric knowledge of LLMs. We selected three variations FT-L~\citep{meng2022locating}, FT-M~\citep{zhang2024comprehensive}, and LoRA~\citep{hu2022lora}, which mitigate the catastrophic forgetting and overfitting issues of standard fine-tuning.
    \item \textbf{In-Context Editing} is a training-free paradigm that associates LLMs with in-context knowledge directly~\citep{zheng-etal-2023-edit,shi2024retrieval,fei2024retrieval}. We adopted a simple baseline ICE method in~\cite{zheng-etal-2023-edit} that puts the new fact in context and does not require demonstrations.
    \item \textbf{Memory-based} methods usually maintain a memory module for knowledge storage and updating. We selected a typical technique GRACE~\citep{hartvigsen2024aging}, which manages a discrete codebook and does not modify the original parameters. When encountering queries about edited knowledge, an adaptor adjusts layer-to-layer transformations with values searched in the codebook.
    
\end{itemize}

\clearpage
\newpage
\section{Results and Analysis}
    \vspace{-0.25cm}
In this section, we comprehensively analyze the experiment results on 9 domains and the overall performance on the whole {\halluedit} for different knowledge editing techniques from five facets including \textit{Efficacy}, \textit{Generalization}, \textit{Portability}, \textit{Locality}, and \textit{Robustness}.

\begin{figure}[t]
    \centering

\includegraphics[width=1\textwidth]{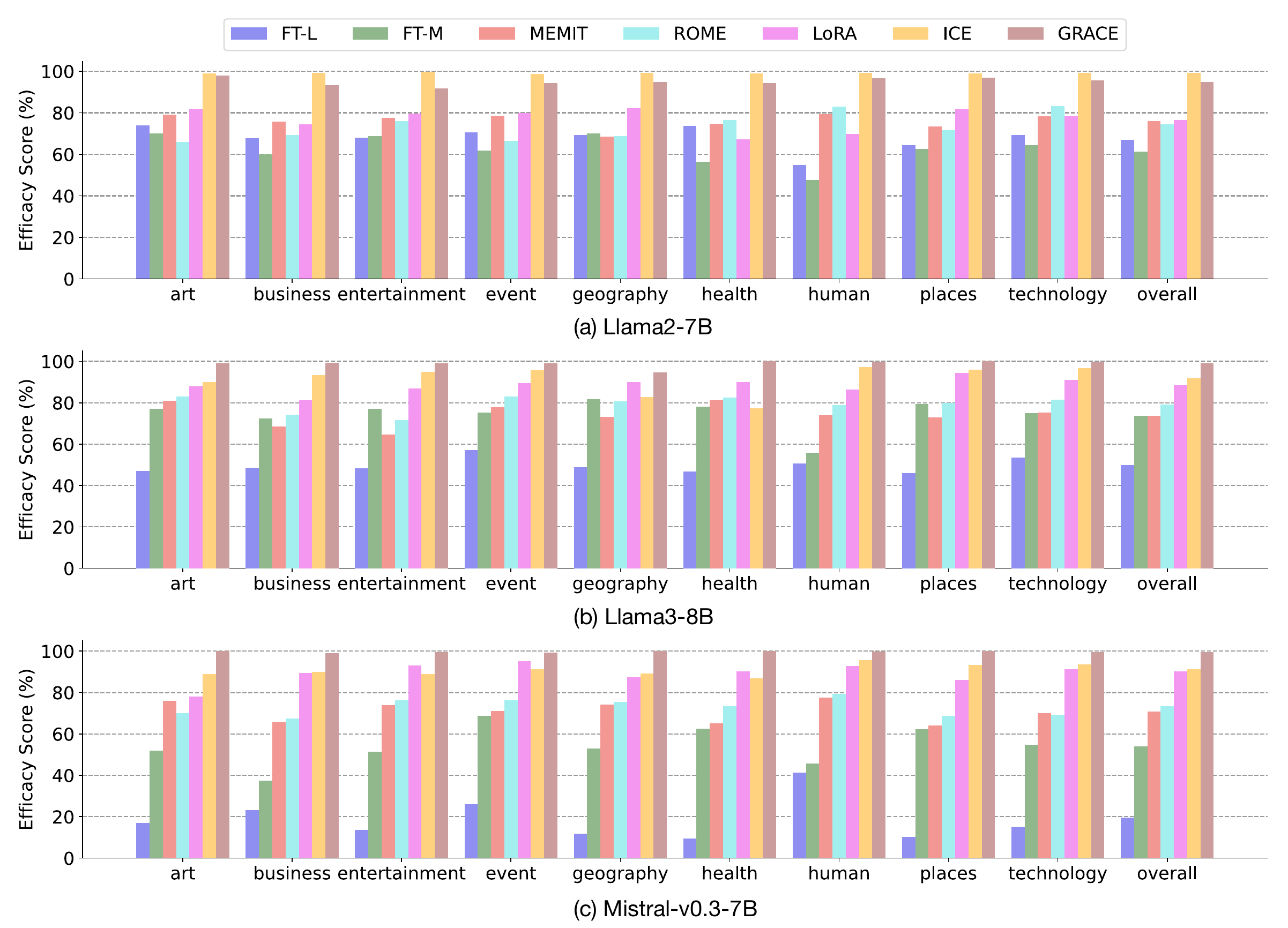}
    \vspace{-0.55cm}
    \caption{\textbf{Efficacy Scores of Knowledge Editing Methods}. The ``overall'' refers to the Efficacy Score (\%) on the whole {\halluedit} embracing 9 domains for different methods. The Efficacy Score on each domain is also reported. Efficacy scores (\%) are measured by the accuracy on Efficacy Evaluation Question-answer Pairs, where the pre-edit scores of each LLM are ensured $0\%$.} 
    \vspace{-0.4cm}    
    \label{fig:efficacy}
\end{figure}

    \vspace{-0.15cm}
\subsection{Facet 1: Efficacy}
    \vspace{-0.05cm}
Since we have ensured that LLMs generate hallucinated answers to the Efficacy Evaluation Questions before editing,  the pre-edit Efficacy Score for all editing  techniques is $0\%$. Thus, Efficacy Scores in Figure~\ref{fig:efficacy} can directly reflect the effectiveness of different  techniques in correcting real-world hallucinations.
We find that \textbf{the effectiveness of some techniques can be far from what their performance on previous datasets suggests}, implying the potential unreliability of their previous evaluation. For example, as shown in Table~\ref{pre-edit Performance}, although FT-M achieves near $100\%$ performance in existing datasets such as $\text{WikiData}_{\text{recent}}$, ZsRE, and WikiBio, its overall Efficacy Scores on Llama2-7B and  Mistral-v0.3-7B are only around $60\%$. There is a similar performance drop for MEMIT.

Second, based on the overall Efficacy Scores across three LLMs, \textbf{the following effectiveness ranking generally holds: FT-L $<$ FT-M $<$ MEMIT $<$ ROME $<$ LoRA $<$ ICE $<$ GRACE}. We can observe that ICE and GRACE, which both preserve original weights in LLMs, outperform the other methods, implying \textbf{the potential disadvantage of directly modifying parameters for knowledge editing}. 

Third, we notice that \textbf{efficacy scores of knowledge editing techniques could highly depend on domains and LLMs}. For example, the scores of FT-L on different domains and LLMs could be highly distinct. The performance of FT-L and FT-M on Llama3-8B is higher than that on Mistral-v0.3-7B.
\begin{center}
\vspace{-2mm}
\begin{tcolorbox}[width=0.99\linewidth, boxrule=3pt, colback=gray!20, colframe=gray!20]
\textbf{Insight 1:} 
(1) The current assessment of knowledge editing could be unreliable;
(2) ICE and GRACE outperform parameter-modifying editing techniques such as fine-tuning  and ``Locate-then-Edit'' methods on \textit{Efficacy};
(3) Domains and LLMs could have a high impact on \textit{Efficacy}.
\end{tcolorbox}
\end{center}

\newpage

\newpage
\subsection{Facet 2: Generalization}
\begin{figure*}[t]
    \centering
    \includegraphics[width=1\textwidth]{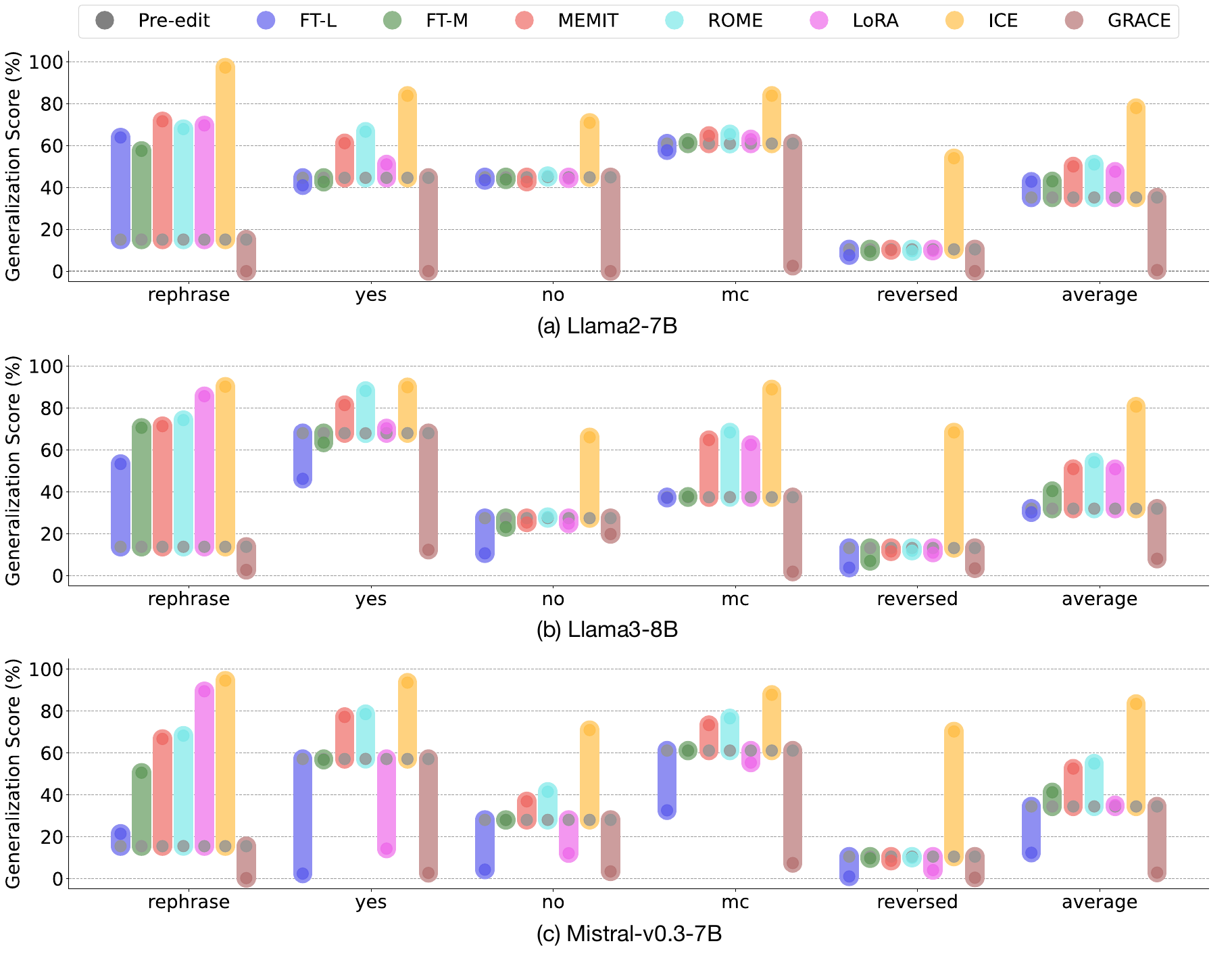}
    \vspace{-4mm}
    \caption{\textbf{Generalization Scores of Knowledge Editing Methods}. Generalization Scores (\%) are measured by accuracy on five types of Generalization Evaluation Questions including Rephrased Questions (``rephrase''), Yes-or-No Questions with ``Yes'' or ``No'' as  answers (``yes'' or ``no''), Multi-Choice Questions (``mc''), Reversed Questions (``reversed''). The ``average'' refers to averaged scores over five question types.
    The figure only shows the overall Generalization Scores for each type on the whole {\halluedit}. Generalization Scores for each domain are given in Appendix~\ref{Generalization Scores of Knowledge Editing Methods on All the Domains}.
    } 
        \vspace{-0.3cm}
    \label{fig:generalization}
\end{figure*}

\vspace{-0.5mm}
As shown in Figure~\ref{fig:generalization}, even though the pre-edit Efficacy Score performances for different editing techniques on three LLMs are ensured $0\%$, it is worth noting that the pre-edit Generalization Score performance is not $0\%$ for each question type, illustrating that \textbf{the manifestation of hallucination actually depends on the design of question prompts}. Given a group of diverse question prompts for the same knowledge triplet, LLMs may hallucinate some questions but answer others correctly.

Surprisingly, we find that \textbf{post-edit Generalization Scores could even be lower than pre-edit scores} for the same LLM and question type, demonstrating the potential negative effect caused by knowledge editing. In more detail, we can observe a clear performance drop for GRACE across all the question types, and for FT-L and LoRA on some question types.

Comparing the ranking of Efficacy Scores in Figure~\ref{fig:efficacy} with Figure~\ref{fig:generalization}, we can explicitly see that \textbf{higher Efficacy Scores do not also necessarily indicate higher Generalization Scores}. Especially, although GRACE almost surpasses all the other editing techniques regarding Efficacy Scores, it largely degrades the Generalization Scores compared to pre-edit performance. In addition,  \textbf{all editing methods except ICE only slightly improve or even hurt Generalization Scores}.

\begin{center}
\vspace{-2mm}
\begin{tcolorbox}[width=0.99\linewidth, boxrule=3pt, colback=gray!20, colframe=gray!20]
\textbf{Insight 2:} 
(1) The manifestation of hallucination depends on question design;
(2) Higher \textit{Efficacy} Scores do not also necessarily indicate higher \textit{Generalization} Scores;
(3) All editing techniques except ICE only slightly improve or negatively impact the \textit{Generalization} performance.
\end{tcolorbox}
\end{center}

\clearpage
\newpage
\subsection{Facet 3: Portability}

\begin{figure*}[t]
    \centering
    \includegraphics[width=1\textwidth]{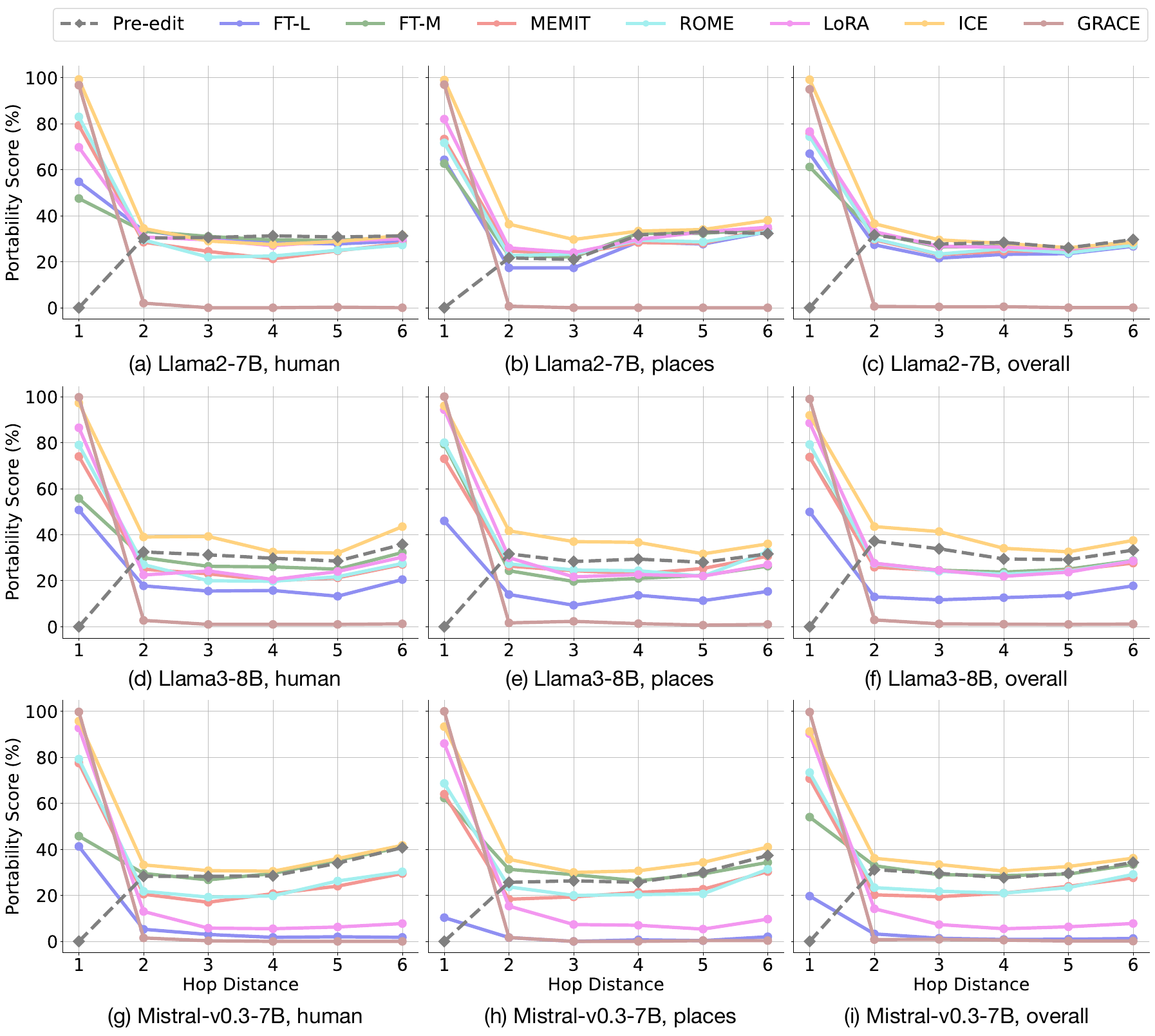}
    \caption{\textbf{Portability Scores of Knowledge Editing Methods}. Portability Scores (\%) are measured by the accuracy on Portability Evaluation Questions, which are Efficacy Evaluation Questions  with $N$ hops ($N = 1 \sim 6$). The Portability Evaluation Questions are the same as Efficacy Evaluation Questions when $N$ is 1. The Portability Scores on two domains ``human'' and ``places'' are  reported in the figure. The results for more domains are given in Appendix~\ref{Portability Scores of Knowledge Editing Methods on All the Domains}.  The ``overall'' refers to the Portability Score (\%) on the whole {\halluedit} embracing 9 domains.} 
        \vspace{-0.4cm}
    \label{fig:portability}
\end{figure*}

Figure~\ref{fig:portability} demonstrates the pre-edit and post-edit Portability Scores for  Portability Evaluation Questions with $N$ hops ($N = 1 \sim 6$). When $N = 1$, the Portability Evaluation Questions are the same as Efficacy Evaluation Questions, suggesting that the Portability Scores are 0. Similar to Figure~\ref{fig:generalization}, we discover that the pre-edit Portability Scores are not zero for $2 \sim 6$ hops, indicating \textbf{LLMs do not necessarily need to reason based on single-hop knowledge to answer multi-hop questions}. We hypothesize that this is because LLMs may directly memorize the answers to multi-hop questions.

We surprisingly find that except that ICE may bring marginal improvement to the pre-edit performance, \textbf{the other knowledge editing techniques even mostly underperform pre-edit Portability Scores}, showing another type of negative effect of knowledge editing and \textbf{LLMs may not really reason with the edited knowledge in multi-hop questions} for most knowledge editing methods. Comparing single-hop and  multi-hop performance, we observe a sharp decrease for all the editing methods, which further underscores \textbf{the challenges of answering multi-hop questions with edited knowledge}. 

\begin{center}
\vspace{-1.5mm}
\begin{tcolorbox}[width=0.99\linewidth, boxrule=3pt, colback=gray!20, colframe=gray!20]
\textbf{Insight 3:} 
(1) LLMs may memorize answers rather than reason based on single-hop knowledge for multi-hop questions;
(2) Editing methods marginally improve or  degrade pre-edit \textit{Portability} Scores, implying LLMs may not really reason with edited knowledge in multi-hop questions.
\end{tcolorbox}
\end{center}
\newpage

\newpage
\subsection{Facet 4: Locality}
\begin{figure*}[t]
    \centering
    \includegraphics[width=1\textwidth]{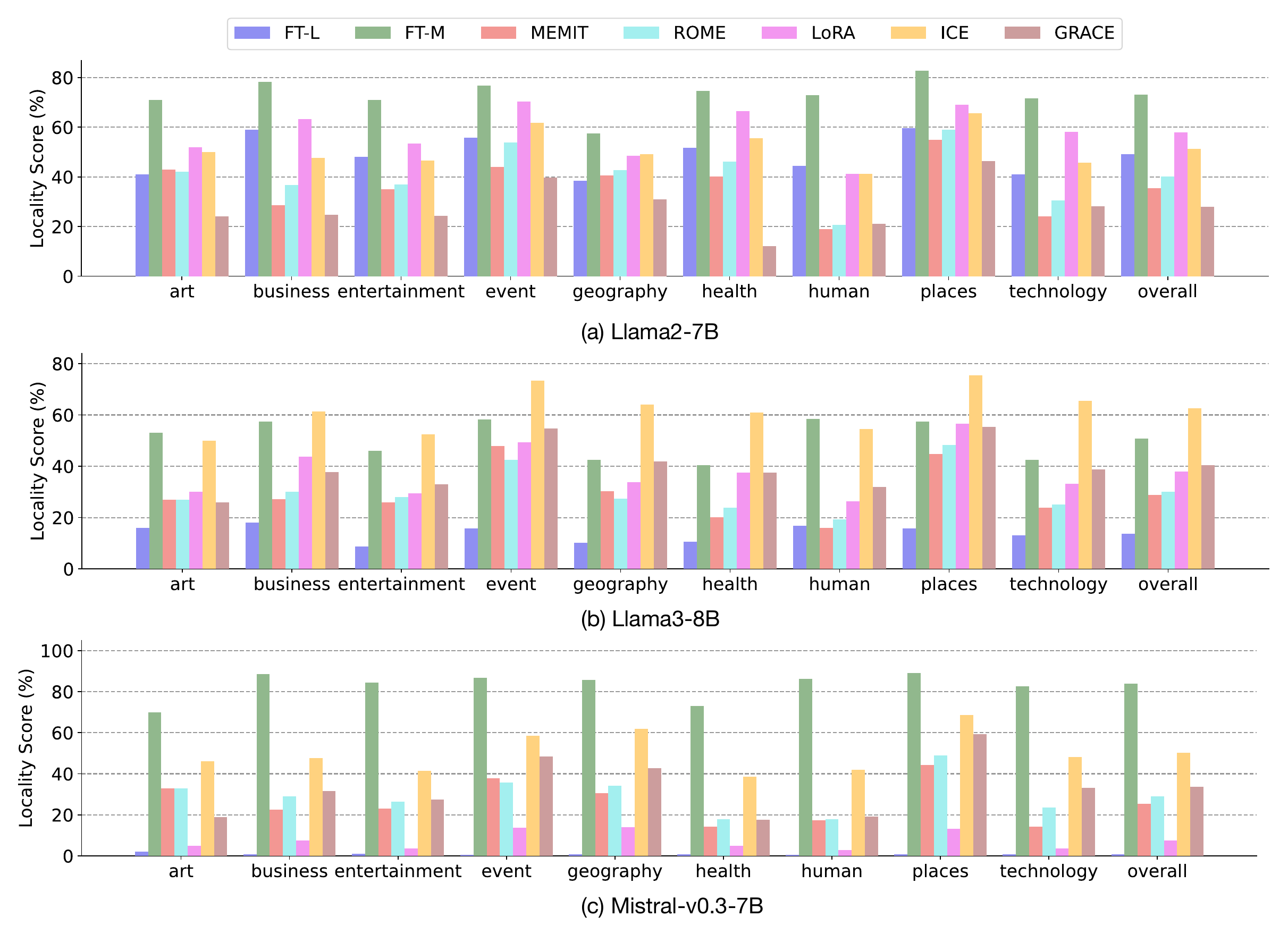}
            \vspace{-0.5cm}
    \caption{\textbf{Locality Scores of Knowledge Editing Methods}. Locality Scores (\%) are measured by the unchanging rate on Locality Evaluation Questions after applying knowledge editing methods on LLMs. A higher Locality Score indicates that there is a higher percentage of LLMs' answers to the unrelated questions keeping the same and a less side effect on general knowledge in LLMs.  The ``overall'' refers to the Locality Score (\%) on the whole {\halluedit} embracing 9 domains for different methods. The Locality Score on each domain is also reported in the figure.} 
        \vspace{-0.37cm}
    \label{fig:Locality}
\end{figure*}

Figure~\ref{fig:Locality} shows the Locality Scores of different editing techniques in each domain and the whole {\halluedit}, reflecting the side effect of knowledge editing on unrelated knowledge encoded in LLMs. Based on the overall Locality Scores, we can observe that \textbf{the performance of all editing methods except FT-M and ICE is unsatisfactory}. In particular, the overall Locality Scores for all editing techniques except FT-M and ICE on Llama3-8B and Mistral-v0.3-7B are below $40\%$, suggesting a high undesired impact on LLMs' answers to unrelated factual questions, though FT-M achieves an overall score of around $80\%$ on Mistral-v0.3-7B and ICE gains $60\%$ on Llama3-8B. 

Furthermore, we notice that \textbf{domains and LLMs have a high impact on the Locality Scores of knowledge editing methods}. For example, the Locality Score for ICE in the ``places'' domain in Llama3-8B is near $80\%$, while the performance drops to only about $50\%$ in the ``art'' domain for the same LLM. Although FT-L obtains a Locality Score around $60\%$ in the ``business'' domain on Llama2-7B, its performance in the same domain on Mistral-v0.3-7B is almost $0\%$.

Due to the impact of LLMs, we observe that \textbf{the rankings by Locality Scores for editing techniques on different LLMs are highly distinct}. For example, the Locality ranking on  Llama2-7B is GRACE $<$ MEMIT $<$ ROME $<$ FT-L $<$ ICE $<$ LoRA $<$ FT-M. However, the ranking changes to FT-L $<$ LoRA $<$ MEMIT $<$ ROME $<$ GRACE $<$ ICE $<$ FT-M on Mistral-v0.3-7B. Comparing Figure~\ref{fig:efficacy} with Figure~\ref{fig:Locality}, we find \textbf{there is no noticeable correlation between Efficacy and Locality for different editing techniques}. FT-M achieves relatively high Locality Scores despite its low Efficacy Scores. 

\begin{center}
\vspace{-1.7mm}
\begin{tcolorbox}[width=0.99\linewidth, boxrule=3pt, colback=gray!20, colframe=gray!20]
\textbf{Insight 4:} 
(1) \textit{Locality} Scores of editing methods except FT-M and ICE are unsatisfactory;
(2) Domains and LLMs have a high impact on \textit{Locality} Scores, and \textit{Locality} rankings are distinct across different LLMs;
(3) \textit{Efficacy} does not have a noticeable correlation with \textit{Locality}.
\end{tcolorbox}
\end{center}

\newpage
\subsection{Facet 5: Robustness}

We proposed Robustness Scores (\%) to evaluate the resistance of edited knowledge against distractions in prompts. Initially ($M = 0$), LLMs are assessed with Efficacy Evaluation Questions. Then ($M = 1 \sim 10$), LLMs are sequentially prompted with Robutness Evaluation Questions, which are exemplified in Figure~\ref{fig:Framework}, for $M$ turns. Robustness Scores are calculated with the percentage of ``Yes'' in each round. A higher Robustness Score indicates that there is a larger percentage of LLMs can resist external manipulations in the prompt and a higher extent of robustness for the edited knowledge.

First, based on overall Robustness Scores, we observe that \textbf{LLMs themselves have a large impact on the robustness of edited knowledge}. \textbf{The same editing method could show distinct trends as turns increase on different LLMs}. For example, all editing methods have a sharp drop when turns go up on Llama2-7B, showing a low level of robustness. However, MEMIT, ROME on Llama3-8B and Mistral-v0.3-7B maintain almost the same and relatively high performance as turns increase, suggesting a comparatively high level of robustness for the edited knowledge.

Then, we notice that \textbf{both ICE and GRACE have a low level of robustness} though they outperform the other five editing techniques regarding Efficacy Scores, showing \textbf{the potential weaknesses on robustness of parameter-preserving knowledge editing methods}. However, parameter-modifying  editing techniques do not necessarily have high robustness, which is exemplified by LoRA.

\begin{figure*}[t]
    \centering
    \includegraphics[width=1\textwidth]{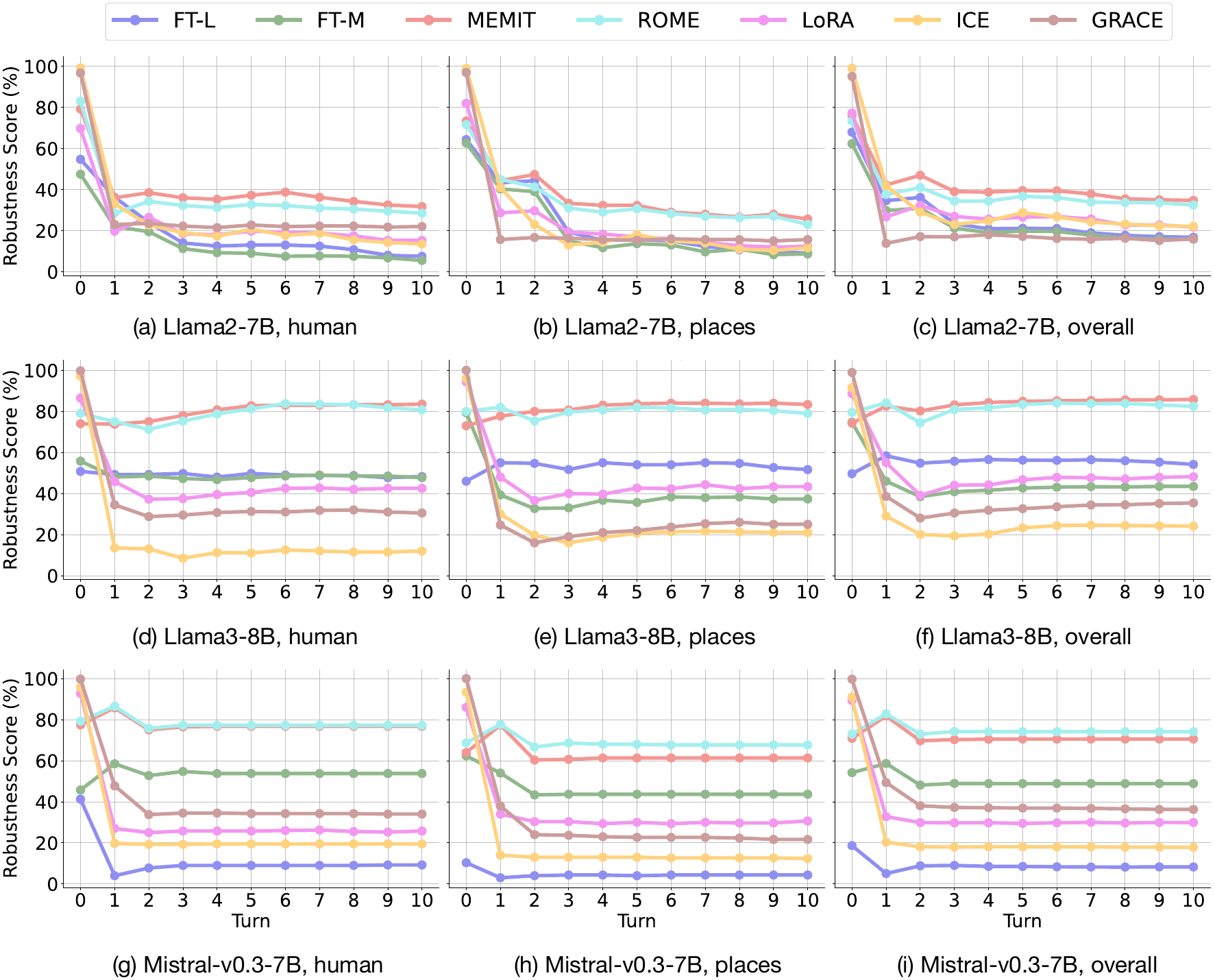}
    \caption{\textbf{Robustness Scores of Knowledge Editing Methods}. Robustness Scores are calculated by the accuracy on Robustness Evaluation Questions with $M$ turns ($M = 1 \sim 10$). We regard Efficacy Scores as the Robustness Scores when $M$ is 0. The Robustness Scores on two domains ``human'' and ``places'' are reported in the figure. The results for more domains are given in Appendix~\ref{Robustness Scores of Knowledge Editing Methods on All the Domains}. The ``overall'' refers to the Robustness Score (\%) on the whole {\halluedit} embracing 9 domains.} 
        \vspace{-0.4cm}
    \label{fig:robutness}
\end{figure*}

\begin{center}
\vspace{-1.5mm}
\begin{tcolorbox}[width=0.99\linewidth, boxrule=3pt, colback=gray!20, colframe=gray!20]
\textbf{Insight 5:} 
(1) LLMs have a large impact on the \textit{Robustness} of edited knowledge;
(2) Parameter-preserving knowledge editing methods such as ICE and GRACE potentially have low \textit{Robustness}.
\end{tcolorbox}
\end{center}
\newpage

\section{Related Work}

Knowledge editing techniques have attracted increasing attention for their efficiency advantages in addressing obsolete or hallucinated information in LLMs~\citep{wang2023knowledge,zhang2024comprehensive}. In general, the existing editing techniques can be categorized into four types including \textit{Locate-then-edit}~\citep{meng2022locating,meng2023massediting}, \textit{Fine-tuning based}~\citep{gangadhar2024model,zhu2020modifying,wang2024roselora}, \textit{In-Context Editing}~\citep{zheng-etal-2023-edit,shi2024retrieval,fei2024retrieval}, and \textit{Memory-based}~\citep{wang2024wise,hartvigsen2024aging,mitchell2022memory,yu2023melo}. Recently, many benchmarks have been built to investigate the properties of knowledge editing from different perspectives~\citep{rosati2024long,wu2023eva,ge2024well,ma2023untying,wei2023assessing,wei2024mlake,zhong2023mquake,lin2024navigating,huang2024kebench,liu2024codeupdatearena,akyurek2023dune,li2024mike,li2024unveiling,li2023evaluating,gu2024model,powell2024taxi,yang2025mirage,du2025mmkebench,zhang2024mc}. For example, \cite{gu2024model} proposed a benchmark to assess the side effect of 4 popular editing methods on 3 LLMs across 8 general capacity tasks. \cite{rosati2024long} built a new evaluation protocol to measure the efficacy and impact of knowledge editing in long-form generation. \cite{wei2024mlake} introduced a multilingual knowledge editing benchmark embracing five languages. However, considering the fundamental motivation of applying knowledge editing to LLMs, which is to correct hallucinations, there is a pressing need to build a real-world hallucination dataset with rigorous verification and systematically analyze the performance of different editing methods. Thus, we proposed {\halluedit} to fill in the gap and provided new insights to facilitate the progress in the field of knowledge editing.

\section{Conclusion}

In this paper, we have built a new benchmark {\halluedit} to holistically assess diverse knowledge editing techniques in correcting real-world hallucinations. First, we meticulously construct a massive and comprehensive hallucination dataset based on Wikidata with $9$ domains, $26$ topics, and more than $6,000$ hallucinations. Then, we systematically investigate the performance of different knowledge editing methods from five perspectives including \textit{Efficacy}, \textit{Generalization}, \textit{Portability}, \textit{Locality}, and \textit{Robustness}. Our findings reveal that previous benchmarks cannot reflect the true effectiveness of knowledge editing methods in correcting real-world hallucinations and current editing methods mostly show limited performance across five dimensions. We also offer valuable and actionable insights to inspire future advancements in knowledge editing for large language models.

\section*{Acknowledgments}
This material is based upon work supported by the U.S. Department of Homeland Security under Grant Award Number 17STQAC00001-07-04, NSF awards (SaTC-2241068, IIS-2506643, and POSE-2346158), a Cisco Research Award, and a Microsoft Accelerate Foundation Models Research Award. The views and conclusions contained in this document are those of the authors and should not be interpreted as necessarily representing the official policies, either expressed or implied, of the U.S. Department of Homeland Security and the National Science Foundation.

\clearpage
\newpage
\bibliography{main_ref}

\begin{thebibliography}{135}
\providecommand{\natexlab}[1]{#1}
\providecommand{\url}[1]{\texttt{#1}}
\expandafter\ifx\csname urlstyle\endcsname\relax
  \providecommand{\doi}[1]{doi: #1}\else
  \providecommand{\doi}{doi: \begingroup \urlstyle{rm}\Url}\fi

\bibitem[Aghajari et~al.(2023)Aghajari, Baumer, and DiFranzo]{aghajari2023reviewing}
Zhila Aghajari, Eric~PS Baumer, and Dominic DiFranzo.
\newblock Reviewing interventions to address misinformation: the need to expand our vision beyond an individualistic focus.
\newblock \textit{Proceedings of the ACM on Human-Computer Interaction}, 7\penalty0 (CSCW1):\penalty0 1--34, 2023.

\bibitem[Aky{\"u}rek et~al.(2023)Aky{\"u}rek, Pan, Kuwanto, and Wijaya]{akyurek2023dune}
Afra~Feyza Aky{\"u}rek, Eric Pan, Garry Kuwanto, and Derry Wijaya.
\newblock Dune: Dataset for unified editing.
\newblock \textit{ArXiv preprint}, abs/2311.16087, 2023.
\newblock URL \url{https://arxiv.org/abs/2311.16087}.

\bibitem[Bak-Coleman et~al.(2022)Bak-Coleman, Kennedy, Wack, Beers, Schafer, Spiro, Starbird, and West]{bak2022combining}
Joseph~B Bak-Coleman, Ian Kennedy, Morgan Wack, Andrew Beers, Joseph~S Schafer, Emma~S Spiro, Kate Starbird, and Jevin~D West.
\newblock Combining interventions to reduce the spread of viral misinformation.
\newblock \textit{Nature Human Behaviour}, 6\penalty0 (10):\penalty0 1372--1380, 2022.

\bibitem[Beigi et~al.(2024)Beigi, Tan, Mudiam, Chen, Shu, and Liu]{beigi2024model}
Alimohammad Beigi, Zhen Tan, Nivedh Mudiam, Canyu Chen, Kai Shu, and Huan Liu.
\newblock Model attribution in machine-generated disinformation: A domain generalization approach with supervised contrastive learning.
\newblock \textit{ArXiv preprint}, abs/2407.21264, 2024.
\newblock URL \url{https://arxiv.org/abs/2407.21264}.

\bibitem[Bi et~al.(2024{\natexlab{a}})Bi, Liu, Mei, Wang, Ji, and Cheng]{bi2024decoding}
Baolong Bi, Shenghua Liu, Lingrui Mei, Yiwei Wang, Pengliang Ji, and Xueqi Cheng.
\newblock Decoding by contrasting knowledge: Enhancing llms' confidence on edited facts.
\newblock \textit{ArXiv preprint}, abs/2405.11613, 2024{\natexlab{a}}.
\newblock URL \url{https://arxiv.org/abs/2405.11613}.

\bibitem[Bi et~al.(2024{\natexlab{b}})Bi, Liu, Wang, Mei, Gao, Fang, and Cheng]{bi2024struedit}
Baolong Bi, Shenghua Liu, Yiwei Wang, Lingrui Mei, Hongcheng Gao, Junfeng Fang, and Xueqi Cheng.
\newblock Struedit: Structured outputs enable the fast and accurate knowledge editing for large language models.
\newblock \textit{ArXiv preprint}, abs/2409.10132, 2024{\natexlab{b}}.
\newblock URL \url{https://arxiv.org/abs/2409.10132}.

\bibitem[Bi et~al.(2024{\natexlab{c}})Bi, Liu, Wang, Mei, Gao, Xu, and Cheng]{bi2024adaptive}
Baolong Bi, Shenghua Liu, Yiwei Wang, Lingrui Mei, Hongcheng Gao, Yilong Xu, and Xueqi Cheng.
\newblock Adaptive token biaser: Knowledge editing via biasing key entities.
\newblock \textit{arXiv preprint arXiv: 2406.12468}, 2024{\natexlab{c}}.

\bibitem[Cai et~al.(2024{\natexlab{a}})Cai, Cao, Guo, Wen, Liu, and Chen]{cai2024editing}
Yuchen Cai, Ding Cao, Rongxi Guo, Yaqin Wen, Guiquan Liu, and Enhong Chen.
\newblock Editing knowledge representation of language lodel via rephrased prefix prompts.
\newblock \textit{ArXiv preprint}, abs/2403.14381, 2024{\natexlab{a}}.
\newblock URL \url{https://arxiv.org/abs/2403.14381}.

\bibitem[Cai et~al.(2024{\natexlab{b}})Cai, Cao, Guo, Wen, Liu, and Chen]{cai2024locating}
Yuchen Cai, Ding Cao, Rongxi Guo, Yaqin Wen, Guiquan Liu, and Enhong Chen.
\newblock Locating and mitigating gender bias in large language models.
\newblock \textit{ArXiv preprint}, abs/2403.14409, 2024{\natexlab{b}}.
\newblock URL \url{https://arxiv.org/abs/2403.14409}.

\bibitem[Chen \& Shu(2024{\natexlab{a}})Chen and Shu]{chen2024combatingmisinformation}
Canyu Chen and Kai Shu.
\newblock Combating misinformation in the age of llms: Opportunities and challenges.
\newblock \textit{AI Magazine}, 2024{\natexlab{a}}.
\newblock \doi{10.1002/aaai.12188}.
\newblock URL \url{https://doi.org/10.1002/aaai.12188}.

\bibitem[Chen \& Shu(2024{\natexlab{b}})Chen and Shu]{chen2024llmgenerated}
Canyu Chen and Kai Shu.
\newblock Can {LLM}-generated misinformation be detected?
\newblock In \textit{The Twelfth International Conference on Learning Representations}, 2024{\natexlab{b}}.
\newblock URL \url{https://openreview.net/forum?id=ccxD4mtkTU}.

\bibitem[Chen et~al.(2022)Chen, Wang, Shapiro, Xiao, Wang, and Shu]{chen2022combating}
Canyu Chen, Haoran Wang, Matthew Shapiro, Yunyu Xiao, Fei Wang, and Kai Shu.
\newblock Combating health misinformation in social media: Characterization, detection, intervention, and open issues.
\newblock \textit{ArXiv preprint}, abs/2211.05289, 2022.
\newblock URL \url{https://arxiv.org/abs/2211.05289}.

\bibitem[Chen et~al.(2024{\natexlab{a}})Chen, Huang, Li, Chen, Lai, Xu, Gu, Gu, Yao, Xiao, Yan, Wang, Torr, Song, and Shu]{chen2024canediting}
Canyu Chen, Baixiang Huang, Zekun Li, Zhaorun Chen, Shiyang Lai, Xiongxiao Xu, Jia-Chen Gu, Jindong Gu, Huaxiu Yao, Chaowei Xiao, Xifeng Yan, William~Yang Wang, Philip Torr, Dawn Song, and Kai Shu.
\newblock Can editing llms inject harm?
\newblock \textit{ArXiv preprint}, abs/2407.20224, 2024{\natexlab{a}}.
\newblock URL \url{https://arxiv.org/abs/2407.20224}.

\bibitem[Chen et~al.(2024{\natexlab{b}})Chen, Wang, Wang, Zhang, Li, and He]{chen2024lifelongKnowledge}
Qizhou Chen, Chengyu Wang, Dakan Wang, Taolin Zhang, Wangyue Li, and Xiaofeng He.
\newblock Lifelong knowledge editing for vision language models with low-rank mixture-of-experts.
\newblock \textit{arXiv preprint arXiv:2411.15432}, 2024{\natexlab{b}}.

\bibitem[Chen et~al.(2024{\natexlab{c}})Chen, Zhang, Li, Huang, Xue, Wang, and He]{chen2024lifelong}
Qizhou Chen, Taolin Zhang, Dongyang Li, Longtao Huang, Hui Xue, Chengyu Wang, and Xiaofeng He.
\newblock Lifelong knowledge editing for llms with retrieval-augmented continuous prompt learning.
\newblock \textit{ArXiv preprint}, abs/2405.03279, 2024{\natexlab{c}}.
\newblock URL \url{https://arxiv.org/abs/2405.03279}.

\bibitem[Chen et~al.(2024{\natexlab{d}})Chen, Cao, Chen, Liu, and Zhao]{chen2024journey}
Yuheng Chen, Pengfei Cao, Yubo Chen, Kang Liu, and Jun Zhao.
\newblock Journey to the center of the knowledge neurons: Discoveries of language-independent knowledge neurons and degenerate knowledge neurons.
\newblock In \textit{Proceedings of the AAAI Conference on Artificial Intelligence}, volume~38, pp.\  17817--17825, 2024{\natexlab{d}}.

\bibitem[Chen et~al.(2024{\natexlab{e}})Chen, Cao, Chen, Liu, and Zhao]{chen2024knowledge}
Yuheng Chen, Pengfei Cao, Yubo Chen, Kang Liu, and Jun Zhao.
\newblock Knowledge localization: Mission not accomplished? enter query localization!
\newblock \textit{ArXiv preprint}, abs/2405.14117, 2024{\natexlab{e}}.
\newblock URL \url{https://arxiv.org/abs/2405.14117}.

\bibitem[Cheng et~al.(2024{\natexlab{a}})Cheng, Ali, Yang, Ling, Zhai, Fei, Xu, Yu, Hu, and Wang]{cheng2024leveraging}
Keyuan Cheng, Muhammad~Asif Ali, Shu Yang, Gang Ling, Yuxuan Zhai, Haoyang Fei, Ke~Xu, Lu~Yu, Lijie Hu, and Di~Wang.
\newblock Leveraging logical rules in knowledge editing: A cherry on the top.
\newblock \textit{ArXiv preprint}, abs/2405.15452, 2024{\natexlab{a}}.
\newblock URL \url{https://arxiv.org/abs/2405.15452}.

\bibitem[Cheng et~al.(2024{\natexlab{b}})Cheng, Lin, Fei, Yu, Ali, Hu, Wang, et~al.]{cheng2024multi}
Keyuan Cheng, Gang Lin, Haoyang Fei, Lu~Yu, Muhammad~Asif Ali, Lijie Hu, Di~Wang, et~al.
\newblock Multi-hop question answering under temporal knowledge editing.
\newblock \textit{ArXiv preprint}, abs/2404.00492, 2024{\natexlab{b}}.
\newblock URL \url{https://arxiv.org/abs/2404.00492}.

\bibitem[Cohen et~al.(2024)Cohen, Biran, Yoran, Globerson, and Geva]{cohen2024evaluating}
Roi Cohen, Eden Biran, Ori Yoran, Amir Globerson, and Mor Geva.
\newblock Evaluating the ripple effects of knowledge editing in language models.
\newblock \textit{Transactions of the Association for Computational Linguistics}, 12:\penalty0 283--298, 2024.

\bibitem[Deng et~al.(2024)Deng, Wei, Pang, Ding, Shen, and Cheng]{deng2024unke}
Jingcheng Deng, Zihao Wei, Liang Pang, Hanxing Ding, Huawei Shen, and Xueqi Cheng.
\newblock Unke: Unstructured knowledge editing in large language models.
\newblock \textit{ArXiv preprint}, abs/2405.15349, 2024.
\newblock URL \url{https://arxiv.org/abs/2405.15349}.

\bibitem[Du. et~al.(2025)Du., Jiang, Gao, Shi, Zheng, Qi, and Li]{du2025mmkebench}
Yuntao Du., Kailin Jiang, Zhi Gao, Chenrui Shi, Zilong Zheng, Siyuan Qi, and Qing Li.
\newblock {MMKE}-bench: A multimodal editing benchmark for diverse visual knowledge.
\newblock In \textit{The Thirteenth International Conference on Learning Representations}, 2025.
\newblock URL \url{https://openreview.net/forum?id=v8qABSeeKO}.

\bibitem[Fang et~al.(2024)Fang, Jiang, Wang, Ma, Wang, He, and Chua]{fang2024alphaedit}
Junfeng Fang, Houcheng Jiang, Kun Wang, Yunshan Ma, Xiang Wang, Xiangnan He, and Tat-seng Chua.
\newblock Alphaedit: Null-space constrained knowledge editing for language models.
\newblock \textit{ArXiv preprint}, abs/2410.02355, 2024.
\newblock URL \url{https://arxiv.org/abs/2410.02355}.

\bibitem[Fei et~al.(2024)Fei, Niu, Xie, Zhang, Bai, Deng, and Han]{fei2024retrieval}
Weizhi Fei, Xueyan Niu, Guoqing Xie, Yanhua Zhang, Bo~Bai, Lei Deng, and Wei Han.
\newblock Retrieval meets reasoning: Dynamic in-context editing for long-text understanding.
\newblock \textit{ArXiv preprint}, abs/2406.12331, 2024.
\newblock URL \url{https://arxiv.org/abs/2406.12331}.

\bibitem[Ferrando et~al.(2024)Ferrando, Sarti, Bisazza, and Costa-juss{\`a}]{ferrando2024primer}
Javier Ferrando, Gabriele Sarti, Arianna Bisazza, and Marta~R Costa-juss{\`a}.
\newblock A primer on the inner workings of transformer-based language models.
\newblock \textit{ArXiv preprint}, abs/2405.00208, 2024.
\newblock URL \url{https://arxiv.org/abs/2405.00208}.

\bibitem[Gangadhar \& Stratos(2024)Gangadhar and Stratos]{gangadhar2024model}
Govind Gangadhar and Karl Stratos.
\newblock Model editing by pure fine-tuning.
\newblock \textit{ArXiv preprint}, abs/2402.11078, 2024.
\newblock URL \url{https://arxiv.org/abs/2402.11078}.

\bibitem[Ge et~al.(2024{\natexlab{a}})Ge, Rudzicz, and Zhu]{ge2024well}
Huaizhi Ge, Frank Rudzicz, and Zining Zhu.
\newblock How well can knowledge edit methods edit perplexing knowledge?
\newblock \textit{ArXiv preprint}, abs/2406.17253, 2024{\natexlab{a}}.
\newblock URL \url{https://arxiv.org/abs/2406.17253}.

\bibitem[Ge et~al.(2024{\natexlab{b}})Ge, Mousavi, Grave, Joulin, Qian, Han, Arefiyan, and Li]{ge2024time}
Xiou Ge, Ali Mousavi, Edouard Grave, Armand Joulin, Kun Qian, Benjamin Han, Mostafa Arefiyan, and Yunyao Li.
\newblock Time sensitive knowledge editing through efficient finetuning.
\newblock \textit{ArXiv preprint}, abs/2406.04496, 2024{\natexlab{b}}.
\newblock URL \url{https://arxiv.org/abs/2406.04496}.

\bibitem[Grimes et~al.(2024)Grimes, Christiani, Shriver, and Connor]{grimes2024concept}
Keltin Grimes, Marco Christiani, David Shriver, and Marissa Connor.
\newblock Concept-rot: Poisoning concepts in large language models with model editing.
\newblock \textit{arXiv preprint arXiv:2412.13341}, 2024.

\bibitem[Gu et~al.(2023)Gu, Zhou, Han, Liu, Wang, and Wang]{gu2023pokemqa}
Hengrui Gu, Kaixiong Zhou, Xiaotian Han, Ninghao Liu, Ruobing Wang, and Xin Wang.
\newblock Pokemqa: Programmable knowledge editing for multi-hop question answering.
\newblock \textit{ArXiv preprint}, abs/2312.15194, 2023.
\newblock URL \url{https://arxiv.org/abs/2312.15194}.

\bibitem[Gu et~al.(2024)Gu, Xu, Ma, Lu, Ling, Chang, and Peng]{gu2024model}
Jia-Chen Gu, Hao-Xiang Xu, Jun-Yu Ma, Pan Lu, Zhen-Hua Ling, Kai-Wei Chang, and Nanyun Peng.
\newblock Model editing harms general abilities of large language models: Regularization to the rescue.
\newblock \textit{ArXiv preprint}, abs/2401.04700, 2024.
\newblock URL \url{https://arxiv.org/abs/2401.04700}.

\bibitem[Gu et~al.(2025)Gu, Chen, Liu, Li, Liu, Tao, Zhang, and Hu]{gu2025editing}
Xiaojie Gu, Guangxu Chen, Shuliang Liu, Jungang Li, Aiwei Liu, Sicheng Tao, Junyan Zhang, and Xuming Hu.
\newblock Editing large language models via adaptive gradient guidance.
\newblock In \textit{AAAI 2025 Workshop on Preventing and Detecting LLM Misinformation (PDLM)}, 2025.
\newblock URL \url{https://openreview.net/forum?id=q8fJI2r1I8}.

\bibitem[Gupta et~al.(2024)Gupta, Rao, and Anumanchipalli]{gupta2024model}
Akshat Gupta, Anurag Rao, and Gopala Anumanchipalli.
\newblock Model editing at scale leads to gradual and catastrophic forgetting.
\newblock \textit{ArXiv preprint}, abs/2401.07453, 2024.
\newblock URL \url{https://arxiv.org/abs/2401.07453}.

\bibitem[Hartvigsen et~al.(2024)Hartvigsen, Sankaranarayanan, Palangi, Kim, and Ghassemi]{hartvigsen2024aging}
Tom Hartvigsen, Swami Sankaranarayanan, Hamid Palangi, Yoon Kim, and Marzyeh Ghassemi.
\newblock Aging with grace: Lifelong model editing with discrete key-value adaptors.
\newblock \textit{Advances in Neural Information Processing Systems}, 36, 2024.

\bibitem[Hartwig et~al.(2024)Hartwig, Doell, and Reuter]{hartwig2024landscape}
Katrin Hartwig, Frederic Doell, and Christian Reuter.
\newblock The landscape of user-centered misinformation interventions-a systematic literature review.
\newblock \textit{ACM Computing Surveys}, 56\penalty0 (11):\penalty0 1--36, 2024.

\bibitem[Hase et~al.(2024{\natexlab{a}})Hase, Bansal, Kim, and Ghandeharioun]{hase2024does}
Peter Hase, Mohit Bansal, Been Kim, and Asma Ghandeharioun.
\newblock Does localization inform editing? surprising differences in causality-based localization vs. knowledge editing in language models.
\newblock \textit{Advances in Neural Information Processing Systems}, 36, 2024{\natexlab{a}}.

\bibitem[Hase et~al.(2024{\natexlab{b}})Hase, Hofweber, Zhou, Stengel-Eskin, and Bansal]{hase2024fundamental}
Peter Hase, Thomas Hofweber, Xiang Zhou, Elias Stengel-Eskin, and Mohit Bansal.
\newblock Fundamental problems with model editing: How should rational belief revision work in llms?
\newblock \textit{ArXiv preprint}, abs/2406.19354, 2024{\natexlab{b}}.
\newblock URL \url{https://arxiv.org/abs/2406.19354}.

\bibitem[He et~al.(2023)He, Ahamad, and Kumar]{he2023reinforcement}
Bing He, Mustaque Ahamad, and Srijan Kumar.
\newblock Reinforcement learning-based counter-misinformation response generation: a case study of covid-19 vaccine misinformation.
\newblock In \textit{Proceedings of the ACM Web Conference 2023}, pp.\  2698--2709, 2023.

\bibitem[Hoelscher-Obermaier et~al.(2023)Hoelscher-Obermaier, Persson, Kran, Konstas, and Barez]{hoelscher2023detecting}
Jason Hoelscher-Obermaier, Julia Persson, Esben Kran, Ioannis Konstas, and Fazl Barez.
\newblock Detecting edit failures in large language models: An improved specificity benchmark.
\newblock \textit{ArXiv preprint}, abs/2305.17553, 2023.
\newblock URL \url{https://arxiv.org/abs/2305.17553}.

\bibitem[Hsueh et~al.(2024)Hsueh, Huang, Lin, Liao, Fang, Huang, and Chen]{hsueh2024editing}
Cheng-Hsun Hsueh, Paul Kuo-Ming Huang, Tzu-Han Lin, Che-Wei Liao, Hung-Chieh Fang, Chao-Wei Huang, and Yun-Nung Chen.
\newblock Editing the mind of giants: An in-depth exploration of pitfalls of knowledge editing in large language models.
\newblock \textit{ArXiv preprint}, abs/2406.01436, 2024.
\newblock URL \url{https://arxiv.org/abs/2406.01436}.

\bibitem[Hu et~al.(2022)Hu, Shen, Wallis, Allen{-}Zhu, Li, Wang, Wang, and Chen]{hu2022lora}
Edward~J. Hu, Yelong Shen, Phillip Wallis, Zeyuan Allen{-}Zhu, Yuanzhi Li, Shean Wang, Lu~Wang, and Weizhu Chen.
\newblock Lora: Low-rank adaptation of large language models.
\newblock In \textit{The Tenth International Conference on Learning Representations, {ICLR} 2022, Virtual Event, April 25-29, 2022}. OpenReview.net, 2022.
\newblock URL \url{https://openreview.net/forum?id=nZeVKeeFYf9}.

\bibitem[Hua et~al.(2024)Hua, Guo, Dong, Zhu, Ng, and Wang]{hua2024propagation}
Wenyue Hua, Jiang Guo, Mingwen Dong, Henghui Zhu, Patrick Ng, and Zhiguo Wang.
\newblock Propagation and pitfalls: Reasoning-based assessment of knowledge editing through counterfactual tasks.
\newblock \textit{ArXiv preprint}, abs/2401.17585, 2024.
\newblock URL \url{https://arxiv.org/abs/2401.17585}.

\bibitem[Huang et~al.(2024{\natexlab{a}})Huang, Chen, and Shu]{huang2024aa_llm}
Baixiang Huang, Canyu Chen, and Kai Shu.
\newblock Authorship attribution in the era of llms: Problems, methodologies, and challenges.
\newblock \textit{ArXiv preprint}, abs/2408.08946, 2024{\natexlab{a}}.
\newblock URL \url{https://arxiv.org/abs/2408.08946}.

\bibitem[Huang et~al.(2024{\natexlab{b}})Huang, Chen, and Shu]{huang2024authorship}
Baixiang Huang, Canyu Chen, and Kai Shu.
\newblock Can large language models identify authorship?, 2024{\natexlab{b}}.
\newblock URL \url{https://arxiv.org/abs/2403.08213}.

\bibitem[Huang et~al.(2024{\natexlab{c}})Huang, Zhong, Liu, Wu, Wang, and Tan]{huang2024kebench}
Han Huang, Haitian Zhong, Qiang Liu, Shu Wu, Liang Wang, and Tieniu Tan.
\newblock Kebench: A benchmark on knowledge editing for large vision-language models.
\newblock \textit{ArXiv preprint}, abs/2403.07350, 2024{\natexlab{c}}.
\newblock URL \url{https://arxiv.org/abs/2403.07350}.

\bibitem[Jiang et~al.(2024{\natexlab{a}})Jiang, Fang, Zhang, Zhang, Wang, Liang, and Wang]{jiang2024neuron}
Houcheng Jiang, Junfeng Fang, Tianyu Zhang, An~Zhang, Ruipeng Wang, Tao Liang, and Xiang Wang.
\newblock Neuron-level sequential editing for large language models.
\newblock \textit{ArXiv preprint}, abs/2410.04045, 2024{\natexlab{a}}.
\newblock URL \url{https://arxiv.org/abs/2410.04045}.

\bibitem[Jiang et~al.(2025)Jiang, Fang, Zhang, Ma, Wan, Wang, He, and Chua]{jiang2025anyedit}
Houcheng Jiang, Junfeng Fang, Ningyu Zhang, Guojun Ma, Mingyang Wan, Xiang Wang, Xiangnan He, and Tat-seng Chua.
\newblock Anyedit: Edit any knowledge encoded in language models.
\newblock \textit{arXiv preprint arXiv:2502.05628}, 2025.

\bibitem[Jiang et~al.(2024{\natexlab{b}})Jiang, Wang, Wu, Zhong, Zeng, Gao, Li, Jiang, Shang, Tang, et~al.]{jiang2024learning}
Yuxin Jiang, Yufei Wang, Chuhan Wu, Wanjun Zhong, Xingshan Zeng, Jiahui Gao, Liangyou Li, Xin Jiang, Lifeng Shang, Ruiming Tang, et~al.
\newblock Learning to edit: Aligning llms with knowledge editing.
\newblock \textit{ArXiv preprint}, abs/2402.11905, 2024{\natexlab{b}}.
\newblock URL \url{https://arxiv.org/abs/2402.11905}.

\bibitem[Li et~al.(2024{\natexlab{a}})Li, Du, Zhang, Chen, Hu, Qi, Jiang, Cheng, and Tian]{li2024mike}
Jiaqi Li, Miaozeng Du, Chuanyi Zhang, Yongrui Chen, Nan Hu, Guilin Qi, Haiyun Jiang, Siyuan Cheng, and Bozhong Tian.
\newblock Mike: A new benchmark for fine-grained multimodal entity knowledge editing.
\newblock \textit{ArXiv preprint}, abs/2402.14835, 2024{\natexlab{a}}.
\newblock URL \url{https://arxiv.org/abs/2402.14835}.

\bibitem[Li et~al.(2024{\natexlab{b}})Li, Deng, Cai, Lu, Chen, and Lam]{li2024consecutive}
Shuaiyi Li, Yang Deng, Deng Cai, Hongyuan Lu, Liang Chen, and Wai Lam.
\newblock Consecutive model editing with batch alongside hook layers.
\newblock \textit{ArXiv preprint}, abs/2403.05330, 2024{\natexlab{b}}.
\newblock URL \url{https://arxiv.org/abs/2403.05330}.

\bibitem[Li et~al.(2024{\natexlab{c}})Li, Li, Song, Yang, Ma, and Yu]{li2024pmet}
Xiaopeng Li, Shasha Li, Shezheng Song, Jing Yang, Jun Ma, and Jie Yu.
\newblock Pmet: Precise model editing in a transformer.
\newblock In \textit{Proceedings of the AAAI Conference on Artificial Intelligence}, volume~38, pp.\  18564--18572, 2024{\natexlab{c}}.

\bibitem[Li et~al.(2024{\natexlab{d}})Li, Wang, Song, Ji, Liu, Li, Ma, and Yu]{li2024identifying}
Xiaopeng Li, Shangwen Wang, Shezheng Song, Bin Ji, Huijun Liu, Shasha Li, Jun Ma, and Jie Yu.
\newblock Identifying knowledge editing types in large language models.
\newblock \textit{arXiv preprint arXiv:2409.19663}, 2024{\natexlab{d}}.

\bibitem[Li et~al.(2024{\natexlab{e}})Li, Fan, Huang, and Li]{li2024learning}
Yanhong Li, Chunling Fan, Mingqing Huang, and Chengming Li.
\newblock Learning from mistakes: A comprehensive review of knowledge editing for large language models.
\newblock In \textit{2024 IEEE International Conference on Smart Internet of Things (SmartIoT)}, pp.\  563--569. IEEE, 2024{\natexlab{e}}.

\bibitem[Li et~al.(2025)Li, Jiang, Chen, Bi, Zhou, Sun, Fang, and Wang]{li2025reinforced}
Zherui Li, Houcheng Jiang, Hao Chen, Baolong Bi, Zhenhong Zhou, Fei Sun, Junfeng Fang, and Xiang Wang.
\newblock Reinforced lifelong editing for language models.
\newblock \textit{arXiv preprint arXiv:2502.05759}, 2025.

\bibitem[Li et~al.(2023{\natexlab{a}})Li, Zhang, Yao, Wang, Chen, and Chen]{li2023unveiling}
Zhoubo Li, Ningyu Zhang, Yunzhi Yao, Mengru Wang, Xi~Chen, and Huajun Chen.
\newblock Unveiling the pitfalls of knowledge editing for large language models.
\newblock \textit{ArXiv preprint}, abs/2310.02129, 2023{\natexlab{a}}.
\newblock URL \url{https://arxiv.org/abs/2310.02129}.

\bibitem[Li et~al.(2024{\natexlab{f}})Li, Zhang, Yao, Wang, Chen, and Chen]{li2024unveiling}
Zhoubo Li, Ningyu Zhang, Yunzhi Yao, Mengru Wang, Xi~Chen, and Huajun Chen.
\newblock Unveiling the pitfalls of knowledge editing for large language models.
\newblock In \textit{The Twelfth International Conference on Learning Representations}, 2024{\natexlab{f}}.
\newblock URL \url{https://openreview.net/forum?id=fNktD3ib16}.

\bibitem[Li et~al.(2023{\natexlab{b}})Li, Arous, Reddy, and Cheung]{li2023evaluating}
Zichao Li, Ines Arous, Siva Reddy, and Jackie Chi~Kit Cheung.
\newblock Evaluating dependencies in fact editing for language models: Specificity and implication awareness.
\newblock In \textit{Findings of the Association for Computational Linguistics: EMNLP 2023}, pp.\  7623--7636, 2023{\natexlab{b}}.

\bibitem[Lin et~al.(2024)Lin, Beigi, Li, Zhou, Zhang, Wang, Yin, and Huang]{lin2024navigating}
Zihao Lin, Mohammad Beigi, Hongxuan Li, Yufan Zhou, Yuxiang Zhang, Qifan Wang, Wenpeng Yin, and Lifu Huang.
\newblock Navigating the dual facets: A comprehensive evaluation of sequential memory editing in large language models.
\newblock \textit{ArXiv preprint}, abs/2402.11122, 2024.
\newblock URL \url{https://arxiv.org/abs/2402.11122}.

\bibitem[Liu et~al.(2024{\natexlab{a}})Liu, Zhang, Liu, Wu, Wu, and Wang]{liu2024uni}
Guofan Liu, Jinghao Zhang, Qiang Liu, Junfei Wu, Shu Wu, and Liang Wang.
\newblock Uni-modal event-agnostic knowledge distillation for multimodal fake news detection.
\newblock \textit{IEEE Transactions on Knowledge and Data Engineering}, 2024{\natexlab{a}}.

\bibitem[Liu et~al.(2024{\natexlab{b}})Liu, Yu, Zhang, Li, Zhang, and Ji]{liu2024evedit}
Jiateng Liu, Pengfei Yu, Yuji Zhang, Sha Li, Zixuan Zhang, and Heng Ji.
\newblock Evedit: Event-based knowledge editing with deductive editing boundaries.
\newblock \textit{ArXiv preprint}, abs/2402.11324, 2024{\natexlab{b}}.
\newblock URL \url{https://arxiv.org/abs/2402.11324}.

\bibitem[Liu et~al.(2025)Liu, Dong, Zhang, Wang, and Gao]{liu2025mitigating}
Tianci Liu, Zihan Dong, Linjun Zhang, Haoyu Wang, and Jing Gao.
\newblock Mitigating heterogeneous token overfitting in llm knowledge editing.
\newblock \textit{arXiv preprint arXiv:2502.00602}, 2025.

\bibitem[Liu et~al.(2024{\natexlab{c}})Liu, Pandit, Ye, Choi, and Durrett]{liu2024codeupdatearena}
Zeyu~Leo Liu, Shrey Pandit, Xi~Ye, Eunsol Choi, and Greg Durrett.
\newblock Codeupdatearena: Benchmarking knowledge editing on api updates.
\newblock \textit{ArXiv preprint}, abs/2407.06249, 2024{\natexlab{c}}.
\newblock URL \url{https://arxiv.org/abs/2407.06249}.

\bibitem[Lu et~al.(2024)Lu, Zhou, Li, Wang, Liu, He, Liu, and Zhang]{lu2024knowledge}
Yifan Lu, Yigeng Zhou, Jing Li, Yequan Wang, Xuebo Liu, Daojing He, Fangming Liu, and Min Zhang.
\newblock Knowledge editing with dynamic knowledge graphs for multi-hop question answering.
\newblock \textit{arXiv preprint arXiv:2412.13782}, 2024.

\bibitem[Ma et~al.(2023)Ma, Gu, Ling, Liu, and Liu]{ma2023untying}
Jun-Yu Ma, Jia-Chen Gu, Zhen-Hua Ling, Quan Liu, and Cong Liu.
\newblock Untying the reversal curse via bidirectional language model editing.
\newblock \textit{ArXiv preprint}, abs/2310.10322, 2023.
\newblock URL \url{https://arxiv.org/abs/2310.10322}.

\bibitem[Meng et~al.(2022)Meng, Bau, Andonian, and Belinkov]{meng2022locating}
Kevin Meng, David Bau, Alex Andonian, and Yonatan Belinkov.
\newblock Locating and editing factual associations in gpt.
\newblock \textit{Advances in Neural Information Processing Systems}, 35:\penalty0 17359--17372, 2022.

\bibitem[Meng et~al.(2023)Meng, Sharma, Andonian, Belinkov, and Bau]{meng2023massediting}
Kevin Meng, Arnab~Sen Sharma, Alex~J Andonian, Yonatan Belinkov, and David Bau.
\newblock Mass-editing memory in a transformer.
\newblock In \textit{The Eleventh International Conference on Learning Representations}, 2023.
\newblock URL \url{https://openreview.net/forum?id=MkbcAHIYgyS}.

\bibitem[Mitchell et~al.(2022)Mitchell, Lin, Bosselut, Manning, and Finn]{mitchell2022memory}
Eric Mitchell, Charles Lin, Antoine Bosselut, Christopher~D. Manning, and Chelsea Finn.
\newblock Memory-based model editing at scale.
\newblock In Kamalika Chaudhuri, Stefanie Jegelka, Le~Song, Csaba Szepesv{\'{a}}ri, Gang Niu, and Sivan Sabato (eds.), \textit{International Conference on Machine Learning, {ICML} 2022, 17-23 July 2022, Baltimore, Maryland, {USA}}, volume 162 of \textit{Proceedings of Machine Learning Research}, pp.\  15817--15831. {PMLR}, 2022.
\newblock URL \url{https://proceedings.mlr.press/v162/mitchell22a.html}.

\bibitem[Nan et~al.(2023)Nan, Sheng, Cao, Zhu, Wang, Yang, Li, and Shu]{nan2023exploiting}
Qiong Nan, Qiang Sheng, Juan Cao, Yongchun Zhu, Danding Wang, Guang Yang, Jintao Li, and Kai Shu.
\newblock Exploiting user comments for early detection of fake news prior to users' commenting.
\newblock \textit{ArXiv preprint}, abs/2310.10429, 2023.
\newblock URL \url{https://arxiv.org/abs/2310.10429}.

\bibitem[Nan et~al.(2024)Nan, Sheng, Cao, Hu, Wang, and Li]{nan2024let}
Qiong Nan, Qiang Sheng, Juan Cao, Beizhe Hu, Danding Wang, and Jintao Li.
\newblock Let silence speak: Enhancing fake news detection with generated comments from large language models.
\newblock In \textit{Proceedings of the 33rd ACM International Conference on Information and Knowledge Management}, pp.\  1732--1742, 2024.

\bibitem[Ni et~al.(2025)Ni, Liu, Wang, Lei, Zhao, Cheng, Zeng, Dong, Xia, Kenthapadi, et~al.]{ni2025towards}
Bo~Ni, Zheyuan Liu, Leyao Wang, Yongjia Lei, Yuying Zhao, Xueqi Cheng, Qingkai Zeng, Luna Dong, Yinglong Xia, Krishnaram Kenthapadi, et~al.
\newblock Towards trustworthy retrieval augmented generation for large language models: A survey.
\newblock \textit{arXiv preprint arXiv:2502.06872}, 2025.

\bibitem[Niu et~al.(2024)Niu, Liu, Zhu, and Penn]{niu2024does}
Jingcheng Niu, Andrew Liu, Zining Zhu, and Gerald Penn.
\newblock What does the knowledge neuron thesis have to do with knowledge?
\newblock \textit{ArXiv preprint}, abs/2405.02421, 2024.
\newblock URL \url{https://arxiv.org/abs/2405.02421}.

\bibitem[Peng et~al.(2024)Peng, Wang, Li, Zeng, Duo, Cao, Hou, and Li]{peng2024event}
Hao Peng, Xiaozhi Wang, Chunyang Li, Kaisheng Zeng, Jiangshan Duo, Yixin Cao, Lei Hou, and Juanzi Li.
\newblock Event-level knowledge editing.
\newblock \textit{ArXiv preprint}, abs/2402.13093, 2024.
\newblock URL \url{https://arxiv.org/abs/2402.13093}.

\bibitem[Powell et~al.(2024)Powell, Gerych, and Hartvigsen]{powell2024taxi}
Derek Powell, Walter Gerych, and Thomas Hartvigsen.
\newblock Taxi: Evaluating categorical knowledge editing for language models.
\newblock \textit{ArXiv preprint}, abs/2404.15004, 2024.
\newblock URL \url{https://arxiv.org/abs/2404.15004}.

\bibitem[Qi et~al.(2024)Qi, Yang, Jiang, Wang, Li, Zhong, Yang, and Zheng]{qi2024context}
Siyuan Qi, Bangcheng Yang, Kailin Jiang, Xiaobo Wang, Jiaqi Li, Yifan Zhong, Yaodong Yang, and Zilong Zheng.
\newblock In-context editing: Learning knowledge from self-induced distributions.
\newblock \textit{ArXiv preprint}, abs/2406.11194, 2024.
\newblock URL \url{https://arxiv.org/abs/2406.11194}.

\bibitem[Rosati et~al.(2024)Rosati, Gonzales, Chen, Yu, Erkan, Kayani, Chavatapalli, Rudzicz, and Sajjad]{rosati2024long}
Domenic Rosati, Robie Gonzales, Jinkun Chen, Xuemin Yu, Melis Erkan, Yahya Kayani, Satya~Deepika Chavatapalli, Frank Rudzicz, and Hassan Sajjad.
\newblock Long-form evaluation of model editing.
\newblock \textit{ArXiv preprint}, abs/2402.09394, 2024.
\newblock URL \url{https://arxiv.org/abs/2402.09394}.

\bibitem[Rozner et~al.(2024)Rozner, Battash, Wolf, and Lindenbaum]{rozner2024knowledge}
Amit Rozner, Barak Battash, Lior Wolf, and Ofir Lindenbaum.
\newblock Knowledge editing in language models via adapted direct preference optimization.
\newblock \textit{arXiv preprint arXiv: 2406.09920}, 2024.

\bibitem[Sharma et~al.(2024{\natexlab{a}})Sharma, Atkinson, and Bau]{sharma2024locating}
Arnab~Sen Sharma, David Atkinson, and David Bau.
\newblock Locating and editing factual associations in mamba.
\newblock \textit{ArXiv preprint}, abs/2404.03646, 2024{\natexlab{a}}.
\newblock URL \url{https://arxiv.org/abs/2404.03646}.

\bibitem[Sharma et~al.(2024{\natexlab{b}})Sharma, Tong, Korbak, Duvenaud, Askell, Bowman, DURMUS, Hatfield-Dodds, Johnston, Kravec, Maxwell, McCandlish, Ndousse, Rausch, Schiefer, Yan, Zhang, and Perez]{sharma2024towards}
Mrinank Sharma, Meg Tong, Tomasz Korbak, David Duvenaud, Amanda Askell, Samuel~R. Bowman, Esin DURMUS, Zac Hatfield-Dodds, Scott~R Johnston, Shauna~M Kravec, Timothy Maxwell, Sam McCandlish, Kamal Ndousse, Oliver Rausch, Nicholas Schiefer, Da~Yan, Miranda Zhang, and Ethan Perez.
\newblock Towards understanding sycophancy in language models.
\newblock In \textit{The Twelfth International Conference on Learning Representations}, 2024{\natexlab{b}}.
\newblock URL \url{https://openreview.net/forum?id=tvhaxkMKAn}.

\bibitem[Shi et~al.(2024)Shi, Tan, Wu, Zhong, Zhou, and Liu]{shi2024retrieval}
Yucheng Shi, Qiaoyu Tan, Xuansheng Wu, Shaochen Zhong, Kaixiong Zhou, and Ninghao Liu.
\newblock Retrieval-enhanced knowledge editing in language models for multi-hop question answering.
\newblock In \textit{Proceedings of the 33rd ACM International Conference on Information and Knowledge Management}, pp.\  2056--2066, 2024.

\bibitem[Shi et~al.(2025)Shi, Yang, Chen, Li, Liu, Li, and Liu]{shi2025searchrag}
Yucheng Shi, Tianze Yang, Canyu Chen, Quanzheng Li, Tianming Liu, Xiang Li, and Ninghao Liu.
\newblock Searchrag: Can search engines be helpful for llm-based medical question answering?
\newblock \textit{arXiv preprint arXiv:2502.13233}, 2025.

\bibitem[Shu et~al.(2017)Shu, Sliva, Wang, Tang, and Liu]{shu2017fake}
Kai Shu, Amy Sliva, Suhang Wang, Jiliang Tang, and Huan Liu.
\newblock Fake news detection on social media: A data mining perspective.
\newblock \textit{ACM SIGKDD explorations newsletter}, 19\penalty0 (1):\penalty0 22--36, 2017.

\bibitem[Solaiman et~al.(2023)Solaiman, Talat, Agnew, Ahmad, Baker, Blodgett, Chen, Daum{\'e}~III, Dodge, Duan, et~al.]{solaiman2023evaluating}
Irene Solaiman, Zeerak Talat, William Agnew, Lama Ahmad, Dylan Baker, Su~Lin Blodgett, Canyu Chen, Hal Daum{\'e}~III, Jesse Dodge, Isabella Duan, et~al.
\newblock Evaluating the social impact of generative ai systems in systems and society.
\newblock \textit{ArXiv preprint}, abs/2306.05949, 2023.
\newblock URL \url{https://arxiv.org/abs/2306.05949}.

\bibitem[Tonmoy et~al.(2024)Tonmoy, Zaman, Jain, Rani, Rawte, Chadha, and Das]{tonmoy2024comprehensive}
SM~Tonmoy, SM~Zaman, Vinija Jain, Anku Rani, Vipula Rawte, Aman Chadha, and Amitava Das.
\newblock A comprehensive survey of hallucination mitigation techniques in large language models.
\newblock \textit{ArXiv preprint}, abs/2401.01313, 2024.
\newblock URL \url{https://arxiv.org/abs/2401.01313}.

\bibitem[Uppaal et~al.(2024)Uppaal, De, He, Zhong, and Hu]{uppaal2024detox}
Rheeya Uppaal, Apratim De, Yiting He, Yiquao Zhong, and Junjie Hu.
\newblock Detox: Toxic subspace projection for model editing.
\newblock \textit{ArXiv preprint}, abs/2405.13967, 2024.
\newblock URL \url{https://arxiv.org/abs/2405.13967}.

\bibitem[Vidgen et~al.(2024)Vidgen, Agrawal, Ahmed, Akinwande, Al-Nuaimi, Alfaraj, Alhajjar, Aroyo, Bavalatti, Blili-Hamelin, et~al.]{vidgen2024introducing}
Bertie Vidgen, Adarsh Agrawal, Ahmed~M Ahmed, Victor Akinwande, Namir Al-Nuaimi, Najla Alfaraj, Elie Alhajjar, Lora Aroyo, Trupti Bavalatti, Borhane Blili-Hamelin, et~al.
\newblock Introducing v0. 5 of the ai safety benchmark from mlcommons.
\newblock \textit{ArXiv preprint}, abs/2404.12241, 2024.
\newblock URL \url{https://arxiv.org/abs/2404.12241}.

\bibitem[Wang et~al.(2023{\natexlab{a}})Wang, Dou, Chen, Sun, Yu, and Shu]{wang2023attacking}
Haoran Wang, Yingtong Dou, Canyu Chen, Lichao Sun, Philip~S Yu, and Kai Shu.
\newblock Attacking fake news detectors via manipulating news social engagement.
\newblock In \textit{Proceedings of the ACM Web Conference 2023}, pp.\  3978--3986, 2023{\natexlab{a}}.

\bibitem[Wang et~al.(2024{\natexlab{a}})Wang, Liu, Zhao, and Gao]{wang2024roselora}
Haoyu Wang, Tianci Liu, Tuo Zhao, and Jing Gao.
\newblock Roselora: Row and column-wise sparse low-rank adaptation of pre-trained language model for knowledge editing and fine-tuning.
\newblock \textit{ArXiv preprint}, abs/2406.10777, 2024{\natexlab{a}}.
\newblock URL \url{https://arxiv.org/abs/2406.10777}.

\bibitem[Wang et~al.(2023{\natexlab{b}})Wang, Liang, Sun, Cao, and Xu]{wang2023cross}
Jiaan Wang, Yunlong Liang, Zengkui Sun, Yuxuan Cao, and Jiarong Xu.
\newblock Cross-lingual knowledge editing in large language models.
\newblock \textit{ArXiv preprint}, abs/2309.08952, 2023{\natexlab{b}}.
\newblock URL \url{https://arxiv.org/abs/2309.08952}.

\bibitem[Wang et~al.(2024{\natexlab{b}})Wang, Yao, Xu, Qiao, Deng, Wang, Chen, Gu, Jiang, Xie, et~al.]{wang2024knowledge}
Mengru Wang, Yunzhi Yao, Ziwen Xu, Shuofei Qiao, Shumin Deng, Peng Wang, Xiang Chen, Jia-Chen Gu, Yong Jiang, Pengjun Xie, et~al.
\newblock Knowledge mechanisms in large language models: A survey and perspective.
\newblock \textit{ArXiv preprint}, abs/2407.15017, 2024{\natexlab{b}}.
\newblock URL \url{https://arxiv.org/abs/2407.15017}.

\bibitem[Wang et~al.(2024{\natexlab{c}})Wang, Zhang, Xu, Xi, Deng, Yao, Zhang, Yang, Wang, and Chen]{wang2024detoxifying}
Mengru Wang, Ningyu Zhang, Ziwen Xu, Zekun Xi, Shumin Deng, Yunzhi Yao, Qishen Zhang, Linyi Yang, Jindong Wang, and Huajun Chen.
\newblock Detoxifying large language models via knowledge editing.
\newblock \textit{ArXiv preprint}, abs/2403.14472, 2024{\natexlab{c}}.
\newblock URL \url{https://arxiv.org/abs/2403.14472}.

\bibitem[Wang et~al.(2024{\natexlab{d}})Wang, Li, Zhang, Xu, Yao, Jiang, Xie, Huang, and Chen]{wang2024wise}
Peng Wang, Zexi Li, Ningyu Zhang, Ziwen Xu, Yunzhi Yao, Yong Jiang, Pengjun Xie, Fei Huang, and Huajun Chen.
\newblock Wise: Rethinking the knowledge memory for lifelong model editing of large language models.
\newblock \textit{ArXiv preprint}, abs/2405.14768, 2024{\natexlab{d}}.
\newblock URL \url{https://arxiv.org/abs/2405.14768}.

\bibitem[Wang \& Li(2024{\natexlab{a}})Wang and Li]{wang2024lemoe}
Renzhi Wang and Piji Li.
\newblock Lemoe: Advanced mixture of experts adaptor for lifelong model editing of large language models.
\newblock \textit{ArXiv preprint}, abs/2406.20030, 2024{\natexlab{a}}.
\newblock URL \url{https://arxiv.org/abs/2406.20030}.

\bibitem[Wang \& Li(2024{\natexlab{b}})Wang and Li]{wang2024semantic}
Renzhi Wang and Piji Li.
\newblock Semantic are beacons: A semantic perspective for unveiling parameter-efficient fine-tuning in knowledge learning.
\newblock \textit{ArXiv preprint}, abs/2405.18292, 2024{\natexlab{b}}.
\newblock URL \url{https://arxiv.org/abs/2405.18292}.

\bibitem[Wang et~al.(2023{\natexlab{c}})Wang, Zhu, Liu, Zheng, Chen, et~al.]{wang2023knowledge}
Song Wang, Yaochen Zhu, Haochen Liu, Zaiyi Zheng, Chen Chen, et~al.
\newblock Knowledge editing for large language models: A survey.
\newblock \textit{ArXiv preprint}, abs/2310.16218, 2023{\natexlab{c}}.
\newblock URL \url{https://arxiv.org/abs/2310.16218}.

\bibitem[Wang et~al.(2024{\natexlab{e}})Wang, Shi, Tu, Yuan, Huang, Jiao, and Lyu]{wang2024earth}
Wenxuan Wang, Juluan Shi, Zhaopeng Tu, Youliang Yuan, Jen-tse Huang, Wenxiang Jiao, and Michael~R Lyu.
\newblock The earth is flat? unveiling factual errors in large language models.
\newblock \textit{ArXiv preprint}, abs/2401.00761, 2024{\natexlab{e}}.
\newblock URL \url{https://arxiv.org/abs/2401.00761}.

\bibitem[Wang et~al.(2024{\natexlab{f}})Wang, Mao, Zhang, Deng, Yao, Shen, Liang, Gu, and Chen]{wang2024editing}
Xiaohan Wang, Shengyu Mao, Ningyu Zhang, Shumin Deng, Yunzhi Yao, Yue Shen, Lei Liang, Jinjie Gu, and Huajun Chen.
\newblock Editing conceptual knowledge for large language models.
\newblock \textit{ArXiv preprint}, abs/2403.06259, 2024{\natexlab{f}}.
\newblock URL \url{https://arxiv.org/abs/2403.06259}.

\bibitem[Wang et~al.(2024{\natexlab{g}})Wang, Chen, Peng, and Chang]{wang2024deepedit}
Yiwei Wang, Muhao Chen, Nanyun Peng, and Kai-Wei Chang.
\newblock Deepedit: Knowledge editing as decoding with constraints.
\newblock \textit{ArXiv preprint}, abs/2401.10471, 2024{\natexlab{g}}.
\newblock URL \url{https://arxiv.org/abs/2401.10471}.

\bibitem[Wei et~al.(2023)Wei, Yu, Ma, Lei, Weng, Song, and Liu]{wei2023assessing}
Yifan Wei, Xiaoyan Yu, Huanhuan Ma, Fangyu Lei, Yixuan Weng, Ran Song, and Kang Liu.
\newblock Assessing knowledge editing in language models via relation perspective.
\newblock \textit{ArXiv preprint}, abs/2311.09053, 2023.
\newblock URL \url{https://arxiv.org/abs/2311.09053}.

\bibitem[Wei et~al.(2024{\natexlab{a}})Wei, Deng, Pang, Ding, Shen, and Cheng]{wei2024mlake}
Zihao Wei, Jingcheng Deng, Liang Pang, Hanxing Ding, Huawei Shen, and Xueqi Cheng.
\newblock Mlake: Multilingual knowledge editing benchmark for large language models.
\newblock \textit{ArXiv preprint}, abs/2404.04990, 2024{\natexlab{a}}.
\newblock URL \url{https://arxiv.org/abs/2404.04990}.

\bibitem[Wei et~al.(2024{\natexlab{b}})Wei, Pang, Ding, Deng, Shen, and Cheng]{wei2024stable}
Zihao Wei, Liang Pang, Hanxing Ding, Jingcheng Deng, Huawei Shen, and Xueqi Cheng.
\newblock Stable knowledge editing in large language models.
\newblock \textit{ArXiv preprint}, abs/2402.13048, 2024{\natexlab{b}}.
\newblock URL \url{https://arxiv.org/abs/2402.13048}.

\bibitem[Wu et~al.(2023)Wu, Peng, Chen, Su, and Sun]{wu2023eva}
Suhang Wu, Minlong Peng, Yue Chen, Jinsong Su, and Mingming Sun.
\newblock Eva-kellm: A new benchmark for evaluating knowledge editing of llms.
\newblock \textit{ArXiv preprint}, abs/2308.09954, 2023.
\newblock URL \url{https://arxiv.org/abs/2308.09954}.

\bibitem[Wu et~al.(2024)Wu, Pan, Wang, and Luu]{wu2024updating}
Xiaobao Wu, Liangming Pan, William~Yang Wang, and Anh~Tuan Luu.
\newblock Updating language models with unstructured facts: Towards practical knowledge editing.
\newblock \textit{ArXiv preprint}, abs/2402.18909, 2024.
\newblock URL \url{https://arxiv.org/abs/2402.18909}.

\bibitem[Wu et~al.(2025)Wu, Ding, Shen, and Tao]{wu2025edit}
Yuchen Wu, Liang Ding, Li~Shen, and Dacheng Tao.
\newblock Edit once, update everywhere: A simple framework for cross-lingual knowledge synchronization in llms.
\newblock \textit{arXiv preprint arXiv:2502.14645}, 2025.

\bibitem[Xie et~al.(2024)Xie, Cao, Chen, Chen, Liu, and Zhao]{xie2024memla}
Jiakuan Xie, Pengfei Cao, Yuheng Chen, Yubo Chen, Kang Liu, and Jun Zhao.
\newblock Memla: Enhancing multilingual knowledge editing with neuron-masked low-rank adaptation.
\newblock \textit{arXiv preprint arXiv: 2406.11566}, 2024.

\bibitem[Xu et~al.(2024)Xu, Zhang, Zhu, Lin, Liu, Wu, Xu, Zhao, Zheng, and Chen]{xu2024editing}
Derong Xu, Ziheng Zhang, Zhihong Zhu, Zhenxi Lin, Qidong Liu, Xian Wu, Tong Xu, Xiangyu Zhao, Yefeng Zheng, and Enhong Chen.
\newblock Editing factual knowledge and explanatory ability of medical large language models.
\newblock \textit{ArXiv preprint}, abs/2402.18099, 2024.
\newblock URL \url{https://arxiv.org/abs/2402.18099}.

\bibitem[Yan et~al.(2024)Yan, Wang, Li, and Zhang]{yan2024potential}
Jianhao Yan, Futing Wang, Yafu Li, and Yue Zhang.
\newblock Potential and challenges of model editing for social debiasing.
\newblock \textit{ArXiv preprint}, abs/2402.13462, 2024.
\newblock URL \url{https://arxiv.org/abs/2402.13462}.

\bibitem[Yang et~al.(2024{\natexlab{a}})Yang, Sun, Ma, Liu, Yin, and Cheng]{yang-etal-2024-butterfly}
Wanli Yang, Fei Sun, Xinyu Ma, Xun Liu, Dawei Yin, and Xueqi Cheng.
\newblock The butterfly effect of model editing: Few edits can trigger large language models collapse.
\newblock In Lun-Wei Ku, Andre Martins, and Vivek Srikumar (eds.), \textit{Findings of the Association for Computational Linguistics: ACL 2024}, pp.\  5419--5437, Bangkok, Thailand, August 2024{\natexlab{a}}. Association for Computational Linguistics.
\newblock \doi{10.18653/v1/2024.findings-acl.322}.
\newblock URL \url{https://aclanthology.org/2024.findings-acl.322/}.

\bibitem[Yang et~al.(2024{\natexlab{b}})Yang, Sun, Tan, Ma, Su, Yin, and Shen]{yang2024fall}
Wanli Yang, Fei Sun, Jiajun Tan, Xinyu Ma, Du~Su, Dawei Yin, and Huawei Shen.
\newblock The fall of rome: Understanding the collapse of llms in model editing.
\newblock In \textit{Findings of the Association for Computational Linguistics: EMNLP 2024}, pp.\  4079--4087, 2024{\natexlab{b}}.

\bibitem[Yang et~al.(2025)Yang, Sun, Tan, Ma, Cao, Yin, Shen, and Cheng]{yang2025mirage}
Wanli Yang, Fei Sun, Jiajun Tan, Xinyu Ma, Qi~Cao, Dawei Yin, Huawei Shen, and Xueqi Cheng.
\newblock The mirage of model editing: Revisiting evaluation in the wild.
\newblock \textit{arXiv preprint arXiv:2502.11177}, 2025.

\bibitem[Yao et~al.(2023)Yao, Wang, Tian, Cheng, Li, Deng, Chen, and Zhang]{yao2023editing}
Yunzhi Yao, Peng Wang, Bozhong Tian, Siyuan Cheng, Zhoubo Li, Shumin Deng, Huajun Chen, and Ningyu Zhang.
\newblock Editing large language models: Problems, methods, and opportunities.
\newblock \textit{ArXiv preprint}, abs/2305.13172, 2023.
\newblock URL \url{https://arxiv.org/abs/2305.13172}.

\bibitem[Yao et~al.(2024)Yao, Zhang, Xi, Wang, Xu, Deng, and Chen]{yao2024knowledge}
Yunzhi Yao, Ningyu Zhang, Zekun Xi, Mengru Wang, Ziwen Xu, Shumin Deng, and Huajun Chen.
\newblock Knowledge circuits in pretrained transformers.
\newblock \textit{ArXiv preprint}, abs/2405.17969, 2024.
\newblock URL \url{https://arxiv.org/abs/2405.17969}.

\bibitem[Yin et~al.(2024)Yin, Jiang, Yang, and Wan]{yin2024history}
Xunjian Yin, Jin Jiang, Liming Yang, and Xiaojun Wan.
\newblock History matters: Temporal knowledge editing in large language model.
\newblock In \textit{Proceedings of the AAAI Conference on Artificial Intelligence}, volume~38, pp.\  19413--19421, 2024.

\bibitem[Youssef et~al.(2024)Youssef, Zhao, Seifert, and Schl{\"o}tterer]{youssef2024has}
Paul Youssef, Zhixue Zhao, Christin Seifert, and J{\"o}rg Schl{\"o}tterer.
\newblock Has this fact been edited? detecting knowledge edits in language models.
\newblock \textit{arXiv preprint arXiv:2405.02765}, 2024.

\bibitem[Youssef et~al.(2025)Youssef, Zhao, Braun, Schl{\"o}tterer, and Seifert]{youssef2025position}
Paul Youssef, Zhixue Zhao, Daniel Braun, J{\"o}rg Schl{\"o}tterer, and Christin Seifert.
\newblock Position: Editing large language models poses serious safety risks.
\newblock \textit{arXiv preprint arXiv:2502.02958}, 2025.

\bibitem[Yu et~al.(2023)Yu, Chen, Zhou, and He]{yu2023melo}
Lang Yu, Qin Chen, Jie Zhou, and Liang He.
\newblock Melo: Enhancing model editing with neuron-indexed dynamic lora.
\newblock \textit{ArXiv preprint}, abs/2312.11795, 2023.
\newblock URL \url{https://arxiv.org/abs/2312.11795}.

\bibitem[Yue et~al.(2024)Yue, Zeng, Lu, Shang, Zhang, and Wang]{yue2024evidence}
Zhenrui Yue, Huimin Zeng, Yimeng Lu, Lanyu Shang, Yang Zhang, and Dong Wang.
\newblock Evidence-driven retrieval augmented response generation for online misinformation.
\newblock In \textit{Proceedings of the 2024 Conference of the North American Chapter of the Association for Computational Linguistics: Human Language Technologies (Volume 1: Long Papers)}, pp.\  5628--5643, 2024.

\bibitem[Zeng et~al.(2024)Zeng, Gu, Yang, Duan, Shi, and Wang]{zeng2024visual}
Zhen Zeng, Leijiang Gu, Xun Yang, Zhangling Duan, Zenglin Shi, and Meng Wang.
\newblock Visual-oriented fine-grained knowledge editing for multimodal large language models.
\newblock \textit{arXiv preprint arXiv:2411.12790}, 2024.

\bibitem[Zhang et~al.(2025{\natexlab{a}})Zhang, Chen, Zheng, Li, and Chen]{zhang2025resolving}
Binchi Zhang, Zhengzhang Chen, Zaiyi Zheng, Jundong Li, and Haifeng Chen.
\newblock Resolving editing-unlearning conflicts: A knowledge codebook framework for large language model updating.
\newblock \textit{arXiv preprint arXiv:2502.00158}, 2025{\natexlab{a}}.

\bibitem[Zhang et~al.(2024{\natexlab{a}})Zhang, Zhang, Yin, Huang, Zhang, Hu, and Wan]{zhang2024mc}
Junzhe Zhang, Huixuan Zhang, Xunjian Yin, Baizhou Huang, Xu~Zhang, Xinyu Hu, and Xiaojun Wan.
\newblock Mc-mke: A fine-grained multimodal knowledge editing benchmark emphasizing modality consistency.
\newblock \textit{arXiv preprint arXiv:2406.13219}, 2024{\natexlab{a}}.

\bibitem[Zhang et~al.(2024{\natexlab{b}})Zhang, Fang, Liu, Ren, Wu, Chen, and Wang]{zhang2024enhancing}
Mengqi Zhang, Bowen Fang, Qiang Liu, Pengjie Ren, Shu Wu, Zhumin Chen, and Liang Wang.
\newblock Enhancing multi-hop reasoning through knowledge erasure in large language model editing.
\newblock \textit{ArXiv preprint}, abs/2408.12456, 2024{\natexlab{b}}.
\newblock URL \url{https://arxiv.org/abs/2408.12456}.

\bibitem[Zhang et~al.(2024{\natexlab{c}})Zhang, Ye, Liu, Ren, Wu, and Chen]{zhang2024knowledge}
Mengqi Zhang, Xiaotian Ye, Qiang Liu, Pengjie Ren, Shu Wu, and Zhumin Chen.
\newblock Knowledge graph enhanced large language model editing.
\newblock \textit{ArXiv preprint}, abs/2402.13593, 2024{\natexlab{c}}.
\newblock URL \url{https://arxiv.org/abs/2402.13593}.

\bibitem[Zhang et~al.(2024{\natexlab{d}})Zhang, Ye, Liu, Ren, Wu, and Chen]{zhang2024uncovering}
Mengqi Zhang, Xiaotian Ye, Qiang Liu, Pengjie Ren, Shu Wu, and Zhumin Chen.
\newblock Uncovering overfitting in large language model editing.
\newblock \textit{ArXiv preprint}, abs/2410.07819, 2024{\natexlab{d}}.
\newblock URL \url{https://arxiv.org/abs/2410.07819}.

\bibitem[Zhang et~al.(2024{\natexlab{e}})Zhang, Xi, Luo, Wang, Tian, Yao, Zhang, Deng, Sun, Liang, et~al.]{zhang2024oneedit}
Ningyu Zhang, Zekun Xi, Yujie Luo, Peng Wang, Bozhong Tian, Yunzhi Yao, Jintian Zhang, Shumin Deng, Mengshu Sun, Lei Liang, et~al.
\newblock Oneedit: A neural-symbolic collaboratively knowledge editing system.
\newblock \textit{ArXiv preprint}, abs/2409.07497, 2024{\natexlab{e}}.
\newblock URL \url{https://arxiv.org/abs/2409.07497}.

\bibitem[Zhang et~al.(2024{\natexlab{f}})Zhang, Yao, Tian, Wang, Deng, Wang, Xi, Mao, Zhang, Ni, et~al.]{zhang2024comprehensive}
Ningyu Zhang, Yunzhi Yao, Bozhong Tian, Peng Wang, Shumin Deng, Mengru Wang, Zekun Xi, Shengyu Mao, Jintian Zhang, Yuansheng Ni, et~al.
\newblock A comprehensive study of knowledge editing for large language models.
\newblock \textit{ArXiv preprint}, abs/2401.01286, 2024{\natexlab{f}}.
\newblock URL \url{https://arxiv.org/abs/2401.01286}.

\bibitem[Zhang et~al.(2024{\natexlab{g}})Zhang, Yu, and Feng]{zhang2024truthx}
Shaolei Zhang, Tian Yu, and Yang Feng.
\newblock Truthx: Alleviating hallucinations by editing large language models in truthful space.
\newblock \textit{ArXiv preprint}, abs/2402.17811, 2024{\natexlab{g}}.
\newblock URL \url{https://arxiv.org/abs/2402.17811}.

\bibitem[Zhang et~al.(2025{\natexlab{b}})Zhang, Fang, Jiang, Bi, Wang, and He]{zhang2025explainable}
Tianyu Zhang, Junfeng Fang, Houcheng Jiang, Baolong Bi, Xiang Wang, and Xiangnan He.
\newblock Explainable and efficient editing for large language models.
\newblock In \textit{THE WEB CONFERENCE 2025}, 2025{\natexlab{b}}.
\newblock URL \url{https://openreview.net/forum?id=iAn7rlIfgc}.

\bibitem[Zhang et~al.(2025{\natexlab{c}})Zhang, Wei, Sun, and Sun]{zhang2025adversarial}
Yihao Zhang, Zeming Wei, Jun Sun, and Meng Sun.
\newblock Adversarial representation engineering: A general model editing framework for large language models.
\newblock \textit{Advances in Neural Information Processing Systems}, 37:\penalty0 126243--126264, 2025{\natexlab{c}}.

\bibitem[Zhang et~al.(2023)Zhang, Li, Cui, Cai, Liu, Fu, Huang, Zhao, Zhang, Chen, Wang, Luu, Bi, Shi, and Shi]{zhang2023hallucination}
Yue Zhang, Yafu Li, Leyang Cui, Deng Cai, Lemao Liu, Tingchen Fu, Xinting Huang, Enbo Zhao, Yu~Zhang, Yulong Chen, Longyue Wang, Anh~Tuan Luu, Wei Bi, Freda Shi, and Shuming Shi.
\newblock Siren's song in the ai ocean: A survey on hallucination in large language models.
\newblock \textit{ArXiv preprint}, abs/2309.01219, 2023.
\newblock URL \url{https://arxiv.org/abs/2309.01219}.

\bibitem[Zhao et~al.(2023)Zhao, Zhou, Li, Tang, Wang, Hou, Min, Zhang, Zhang, Dong, Du, Yang, Chen, Chen, Jiang, Ren, Li, Tang, Liu, Liu, Nie, and Wen]{LLMSurvey}
Wayne~Xin Zhao, Kun Zhou, Junyi Li, Tianyi Tang, Xiaolei Wang, Yupeng Hou, Yingqian Min, Beichen Zhang, Junjie Zhang, Zican Dong, Yifan Du, Chen Yang, Yushuo Chen, Zhipeng Chen, Jinhao Jiang, Ruiyang Ren, Yifan Li, Xinyu Tang, Zikang Liu, Peiyu Liu, Jian-Yun Nie, and Ji-Rong Wen.
\newblock A survey of large language models.
\newblock \textit{ArXiv preprint}, abs/2303.18223, 2023.
\newblock URL \url{https://arxiv.org/abs/2303.18223}.

\bibitem[Zhao et~al.(2025)Zhao, Xu, Li, Wei, and Zhong]{zhao2025fleke}
Zongkai Zhao, Guozeng Xu, Xiuhua Li, Kaiwen Wei, and Jiang Zhong.
\newblock Fleke: Federated locate-then-edit knowledge editing.
\newblock \textit{arXiv preprint arXiv:2502.15677}, 2025.

\bibitem[Zheng et~al.(2023)Zheng, Li, Dong, Fan, Wu, Xu, and Chang]{zheng-etal-2023-edit}
Ce~Zheng, Lei Li, Qingxiu Dong, Yuxuan Fan, Zhiyong Wu, Jingjing Xu, and Baobao Chang.
\newblock Can we edit factual knowledge by in-context learning?
\newblock In Houda Bouamor, Juan Pino, and Kalika Bali (eds.), \textit{Proceedings of the 2023 Conference on Empirical Methods in Natural Language Processing}, pp.\  4862--4876, Singapore, 2023. Association for Computational Linguistics.
\newblock \doi{10.18653/v1/2023.emnlp-main.296}.
\newblock URL \url{https://aclanthology.org/2023.emnlp-main.296}.

\bibitem[Zhong et~al.(2023)Zhong, Wu, Manning, Potts, and Chen]{zhong2023mquake}
Zexuan Zhong, Zhengxuan Wu, Christopher~D Manning, Christopher Potts, and Danqi Chen.
\newblock Mquake: Assessing knowledge editing in language models via multi-hop questions.
\newblock \textit{ArXiv preprint}, abs/2305.14795, 2023.
\newblock URL \url{https://arxiv.org/abs/2305.14795}.

\bibitem[Zhou et~al.(2024)Zhou, Liu, Li, Jin, Qian, Liu, Li, Dou, Ho, and Yu]{zhou2024trustworthiness}
Yujia Zhou, Yan Liu, Xiaoxi Li, Jiajie Jin, Hongjin Qian, Zheng Liu, Chaozhuo Li, Zhicheng Dou, Tsung-Yi Ho, and Philip~S Yu.
\newblock Trustworthiness in retrieval-augmented generation systems: A survey.
\newblock \textit{arXiv preprint arXiv:2409.10102}, 2024.

\bibitem[Zhu et~al.(2020)Zhu, Rawat, Zaheer, Bhojanapalli, Li, Yu, and Kumar]{zhu2020modifying}
Chen Zhu, Ankit~Singh Rawat, Manzil Zaheer, Srinadh Bhojanapalli, Daliang Li, Felix Yu, and Sanjiv Kumar.
\newblock Modifying memories in transformer models.
\newblock \textit{ArXiv preprint}, abs/2012.00363, 2020.
\newblock URL \url{https://arxiv.org/abs/2012.00363}.

\bibitem[Zou et~al.(2023)Zou, Phan, Chen, Campbell, Guo, Ren, Pan, Yin, Mazeika, Dombrowski, et~al.]{zou2023representation}
Andy Zou, Long Phan, Sarah Chen, James Campbell, Phillip Guo, Richard Ren, Alexander Pan, Xuwang Yin, Mantas Mazeika, Ann-Kathrin Dombrowski, et~al.
\newblock Representation engineering: A top-down approach to ai transparency.
\newblock \textit{ArXiv preprint}, abs/2310.01405, 2023.
\newblock URL \url{https://arxiv.org/abs/2310.01405}.

\end{thebibliography}
\bibliographystyle{iclr2025_conference}

\clearpage
\newpage

\newpage

\begin{center}

\LARGE{\textbf{Content of Appendix}}
\end{center}

{
\hypersetup{linktoc=page}
\startcontents[sections]
\printcontents[sections]{l}{1}{\setcounter{tocdepth}{2}}
}

\newpage

\appendix

\newpage

\section{Reproducibility Statement}
\label{Reproducibility Statement}
\vspace{-0.2cm}
We conduct the experiments on NVIDIA RTX A6000 GPUs. The decoding temperatures are $0$ to ensure the reproducibility. The model checkpoints are downloaded from \texttt{\url{https://huggingface.co/}}. The specific download links are as follows:
\vspace{-0.2cm}
\begin{itemize}[leftmargin=*]
    \item Llama2-7B: \url{https://huggingface.co/meta-llama/Llama-2-7b-chat-hf}
    \item Llama3-8B: \url{https://huggingface.co/meta-llama/Meta-Llama-3-8B-Instruct}
    \item Mistral-v0.3-7B: \url{https://huggingface.co/mistralai/Mistral-7B-Instruct-v0.3}
\end{itemize}

\vspace{-0.2cm}
We adopt GPT-4o with the prompt below to generate \textit{Generalization} and \textit{Locality} evaluation questions:

\begin{table*}[h]
\vspace{-0.2cm}
    \centering
    \label{tab:Hallucinated_News_Generation}
    \resizebox{1.0\textwidth}{!}{
    \begin{tabular}{p{14.5cm}}
        \toprule
Given a fact triplet (subject, relation, object), a question asking for the object, and a wrong answer, the correct answer to the question should be the object in the triplet. 
\\\\
Generate the following types of questions:\\
\quad1. Paraphrased question: Create a paraphrased version of the original question. The correct answer should still be the object from the triplet.\\
\quad2. Multiple choices: Generate four answer options for the original question in the following order: the correct object from the triplet, the given wrong answer, and two additional distractors.\\
\quad3. Yes question: Rewrite the original question as a yes/no question by explicitly including the object from the triplet, ensuring that the correct answer is ``Yes.''\\
\quad4. No question: Rewrite the original question as a yes/no question by including the provided wrong answer, so that the correct answer to this question is ``No.''\\
\quad5. Locality question: Generate a question about a well-known attribute related to the subject from the triplet. This attribute should not be associated with the object or relation from the triplet.\\
\quad6. Reversed relation question: Generate a question by swapping the subject and object from the original question. The answer should now be the subject from the triplet.\\\\
Output the result in JSON format with the following keys: ``paraphrased\_question'', ``multiple\_choices'', ``yes\_question'', ``no\_question'', ``locality\_question'', and ``reversed\_relation\_question.''\\
        \bottomrule
    \end{tabular}
    }
    \vspace{-0.2cm}
\end{table*}
We adopt GPT-4o with the following prompt to generate evaluation questions in  \textit{Portability} aspect.
\begin{table*}[h]
    \vspace{-0.2cm}
\label{tab:Hallucinated_News_Generation}
    \resizebox{1.0\textwidth}{!}{
    \begin{tabular}{p{14.5cm}}
        \toprule
        Given a subject, a relation, a 1-hop question, and its answer, create 2-hop, 3-hop, 4-hop, 5-hop, and 6-hop questions, along with their correct answers. \\Always use the provided subject and relation to create multi-hop questions and include the preceding question in the subsequent question (for example, include the 2-hop question in 3-hop question, include the 3-hop question in 4-hop question). \\
        DO NOT include the correct answer to any previous multi-hop question in subsequent ones (for example, do not include the correct answer to the 2-hop question in the 3-hop or 4-hop questions). \\Ensure that the answers for all multi-hop questions are accurate, and do not use 'N/A' as an answer. \\You must include the given subject and relation in all of the 2-hop, 3-hop, 4-hop, 5-hop, and 6-hop questions. Output in JSON format. An example is provided below:\\\\

        Example input:\\
        subject: Amazon, relation: founder\\
        1hop\_question: Who is the Amazon founder? 1hop\_answer: Jeff Bezos\\\\
        
        Example output:\\
        \{\\
        \quad``2hop\_question'': ``Who is the spouse of the Amazon founder?'',
        \quad``2hop\_answer'': ``MacKenzie Scott'',\\
        \quad``3hop\_question'': \quad``Which university did the spouse of the Amazon founder attend for their undergraduate studies?'',
        \quad``3hop\_answer'': \quad``Princeton University'',\\
        \quad``4hop\_question'': \quad``In which city is the university that the spouse of the Amazon founder attended located?'',
        \quad``4hop\_answer'': \quad``Princeton'',\\
        \quad``5hop\_question'': \quad``In which state is the city located where the university that the spouse of the Amazon founder attended is situated?'',
        \quad``5hop\_answer'': \quad``New Jersey'',\\
        \quad``6hop\_question'': \quad``In which country is the state located where the city is situated that contains the university the spouse of the Amazon founder attended?'',
        \quad``6hop\_answer'': \quad``United States'',\\
        \}\\
        \bottomrule
    \end{tabular}
    }
\end{table*}

\clearpage
\newpage
\section{Details of the Benchmarked Knowledge Editing Techniques}
\label{Details of Benchmarked Methods}
\textbf{FT-L}~\citep{zhu2020modifying,meng2022locating}
Constrained Fine-Tuning (FT-L) is a targeted approach to fine-tuning that focuses on adjusting a specific layer within a model's feed-forward network (FFN). Guided by causal tracing results from ROME, FT-L modifies the layer most associated with the desired changes. The goal of FT-L is to fine-tune the model by maximizing the likelihood of the target sequence, particularly focusing on the prediction of the last token, ensuring that the model adapts to modified facts without affecting its broader performance. To achieve this, explicit parameter-space norm constraints are applied to the weights, ensuring minimal interference with unmodified facts and preserving the integrity of the model's original knowledge.
\vspace{0.1cm}

\textbf{FT-M~}\citep{zhang2024comprehensive}
In contrast to FT-L, which fine-tunes by maximizing the probability of all tokens in the target sequence based on the last token's prediction, Fine-Tuning with Masking (FT-M) refines this approach to align more closely with the traditional fine-tuning objective. FT-M also targets the same FFN layer identified by causal tracing but employs a masked training strategy. Specifically, it uses cross-entropy loss on the target answer while masking out the original text, ensuring that the model is trained directly on the relevant target content. This approach mitigates potential deviations from the original fine-tuning objective and provides a more precise adjustment of the model's weights with minimal disruption to unrelated model behavior.
\vspace{0.1cm}

\textbf{LoRA}~\citep{hu2022lora}
Low-Rank Adaptation (LoRA) is a parameter-efficient fine-tuning method that enhances training efficiency by introducing trainable rank decomposition matrices into Transformer layers. Rather than updating the original model parameters directly, LoRA focuses on training expansion and reduction matrices with low intrinsic rank, which allows for significant dimensionality reduction and thus faster training. Specifically, LoRA freezes the pretrained model weights and optimizes rank decomposition matrices to indirectly adapt dense layers without altering the original parameters. This approach greatly reduces the number of trainable parameters needed for downstream tasks, enabling more efficient training and lowering hardware requirements.
\vspace{0.1cm}

\textbf{ROME}~\citep{meng2022locating}
Rank-One Model Editing (ROME) is a ``Locate-then-Edit'' technique designed to modify factual associations within transformer models. ROME localizes these associations along three key dimensions: (1) the MLP module parameters, (2) within a range of middle layers, and (3) specifically during the processing of the last token of the subject. It employs causal intervention to trace the causal effects of hidden state activations, identifying the specific modules that mediate the recall of factual information. Once these decisive MLP modules are localized, ROME makes small, targeted rank-one changes to the parameters of a single MLP module, effectively altering individual factual associations while minimizing disruption to the overall model behavior. This precise parameter adjustment enables direct updates to the model's factual knowledge.
\vspace{0.1cm}

\textbf{MEMIT}~\citep{meng2023massediting}
Mass Editing Memory in a Transformer (MEMIT) builds upon ROME to generalize the editing of feedforward networks (FFNs) in pre-trained transformer models for mass knowledge updates. While ROME focuses on localizing and modifying factual associations within single layers, MEMIT extends this strategy to perform mass edits across a range of critical layers. MEMIT uses causal tracing to identify MLP layers that act as mediators of factual recall, similarly to ROME, but scales the process to enable the simultaneous insertion of thousands of new memories. By explicitly calculating parameter updates, MEMIT targets these critical layers and updates them efficiently, offering a scalable multi-layer update algorithm that enhances and expands upon ROME's capability to modify knowledge across many memories concurrently, achieving orders of magnitude greater scalability.
\vspace{0.1cm}

\textbf{ICE}~\citep{zheng-etal-2023-edit}
In-Context Knowledge Editing (IKE) leverages in-context learning (ICL) to modify model outputs without altering the model's parameters. This approach reduces computational overhead and avoids potential side effects from parameter updates, offering a more efficient and safer way to modify knowledge in large language models. IKE enhances interpretability, providing a human-understandable method for calibrating model behaviors. It achieves this by constructing three types of demonstrations-copy, update, and retain-that guide the model in producing reliable fact editing through the use of a demonstration store. This store, built from training examples, allows the model to retrieve the most relevant demonstrations to inform its responses, improving accuracy in modifying specific factual outputs. In-Context Editing (ICE) is a simple baseline variant of IKE, which  directly uses the new fact as context without additional demonstrations.

\textbf{GRACE}~\citep{hartvigsen2024aging}
GRACE is a knowledge editing method designed to enable thousands of sequential edits without the pitfalls of overfitting or loss of previously learned knowledge, which are common in conventional knowledge editing approaches. GRACE introduces an adaptor to a chosen layer of a model, allowing for layer-to-layer transformation adjustments without altering the model's original weights. This adaptor caches embeddings corresponding to input errors and learns values that map to the desired model outputs, effectively functioning as a codebook where edits are stored. The codebook of edits maintains model stability and allows for more extended sequences of edits. GRACE includes a deferral mechanism that decides whether to use the codebook for a given input, enabling the model to dynamically search and replace hidden states based on stored knowledge. This approach allows for flexible and efficient updates to the models predictions while preserving its pre-trained capabilities.

\vspace{-0.1cm}
\section{A More Detailed Related Work}
\vspace{-0.1cm}

Knowledge Editing has been adopted as one of the mainstream paradigms to address the hallucinations in LLMs efficiently~\citep{chen2024combatingmisinformation,tonmoy2024comprehensive,li2024learning}.
Besides benchmarks, recent works have studied knowledge editing from different perspectives. The first line of works aims to probe into the relationship between localization and editing and gain a deeper understanding of the working mechanisms of different techniques~\citep{wang2024knowledge,niu2024does,hase2024does,hase2024fundamental,ferrando2024primer,gupta2024model,chen2024knowledge,chen2024journey,zou2023representation,yao2024knowledge,wu2025edit}. For example, \cite{hase2024does} found that \textit{Causal Tracing} actually does not provide any insight into which MLP layer is the best option to edit. The second line of works intends to enhance the performance and applicability of knowledge editing in specific scenarios~\citep{rozner2024knowledge,jiang2024neuron,jiang2024learning,zhang2024uncovering,zhang2024knowledge,zhang2024oneedit,zhang2024enhancing,zhang2024truthx,zhang2025resolving,zhang2025explainable,wu2024updating,qi2024context,sharma2024locating,li2024pmet,li2024consecutive,fang2024alphaedit,wang2024lemoe,wang2024semantic,wang2024deepedit,wang2024editing,wang2024wise,wang2023cross,cheng2024multi,cheng2024leveraging,xie2024memla,bi2024adaptive,bi2024struedit,bi2024decoding,chen2024lifelong,chen2024lifelongKnowledge,wei2024stable,fei2024retrieval,xu2024editing,gu2023pokemqa,yin2024history,cai2024editing,liu2024evedit,liu2025mitigating,ge2024time,deng2024unke,peng2024event,zhao2025fleke,jiang2025anyedit,li2025reinforced,lu2024knowledge,zeng2024visual,gu2025editing}. For example, \cite{ma2023untying} proposed a new method
named Bidirectionally Inversible Relationship Modeling (BIRD) to mitigate the \textit{reversal curse} issue in bidirectional language model editing and improve the performance. The third line of works investigates the side effect of knowledge editing techniques~\citep{hsueh2024editing,gu2024model,hoelscher2023detecting,hua2024propagation,yang-etal-2024-butterfly,yang2024fall,li2023unveiling,cohen2024evaluating}. For example, \cite{yang-etal-2024-butterfly} discovered that even one single edit could cause a significant performance degradation in mainstream benchmarks. The fourth line of works explores the potential misuse risks of knowledge editing or its applications beyond correcting hallucinations~\citep{chen2024canediting,uppaal2024detox,wang2024detoxifying,cai2024locating,yan2024potential,zhang2025adversarial,grimes2024concept,li2024identifying,youssef2024has,youssef2025position}. For example, \cite{chen2024canediting} proposed to reformulate knowledge editing as a new type of safety threat, namely \textit{Editing Attack}, and validated its risk of injecting misinformation or bias into LLMs stealthily, suggesting the feasibility of disseminating misinformation or bias  with LLMs as new channels. The social impact of knowledge editing techniques, especially on safety aspect, is worth more attention~\citep{solaiman2023evaluating,vidgen2024introducing}.

\vspace{-0.1cm}
\section{Impact Statement}
\vspace{-0.1cm}

Misinformation is a longstanding threat for online safety and public trust~\citep{chen2022combating,wang2023attacking}. The conventional countermeasures include \textit{detection}~\citep{shu2017fake,nan2024let,nan2023exploiting,liu2024uni}, \textit{intervention}~\citep{bak2022combining,aghajari2023reviewing,hartwig2024landscape,yue2024evidence,he2023reinforcement} and \textit{attribution}~\citep{huang2024aa_llm,huang2024authorship,beigi2024model}. Hallucinations, which could be defined as the non-factual information unintentionally generated by LLMs when used by normal users~\citep{chen2024combatingmisinformation,chen2024llmgenerated}, have become an new type of misinformation and may cause severe information pollution to the online space. Besides methods such as Retrieval-Augmented Generation~\citep{shi2025searchrag,ni2025towards,zhou2024trustworthiness}, knowledge editing is a promising paradigm to correct hallucinations and contribute to the fight against the misinformation crisis in the era of LLMs, due to its advantage of avoiding retraining from scratch. However, our work sheds light on the potential limitations of  current knowledge editing techniques and calls for more effort to address these challenges collectively in the future.

\clearpage
\newpage
\section{More Experiment Results}

\subsection{Generalization Scores of Knowledge Editing Methods on Each Domain}
\label{Generalization Scores of Knowledge Editing Methods on All the Domains}

\begin{figure*}[h]
    \centering
    \includegraphics[width=1\textwidth]{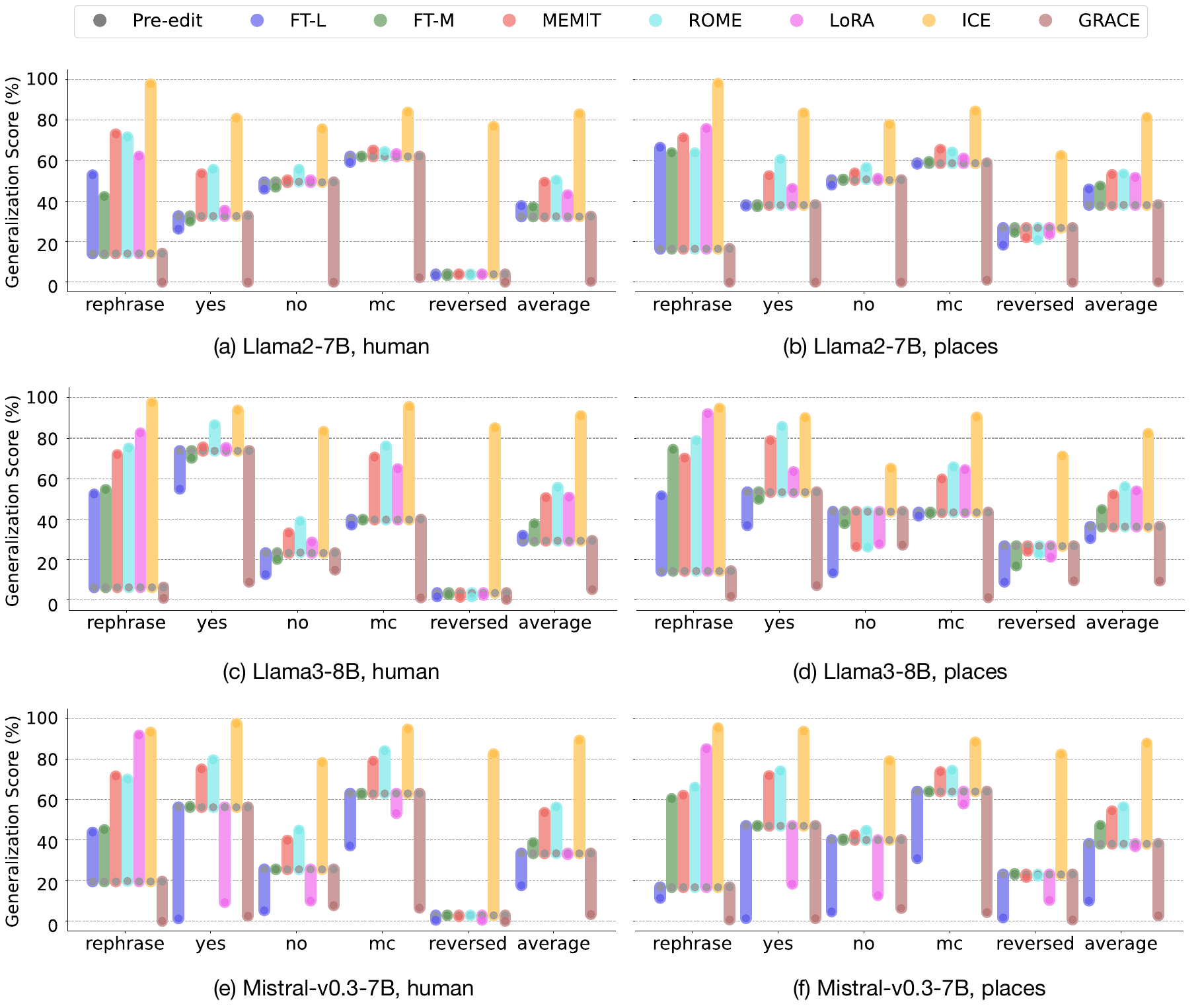}
    \caption{\textbf{Generalization Scores of Knowledge Editing Methods on 3 LLMs and 2 Domains}. Generalization Scores (\%) are measured by the accuracy on five types of Generalization Evaluation Question-answer Pairs including Rephrased Questions (``rephrase''), two types of Yes-or-No Questions with Yes or No as  answers (``yes'' or ``no''), Multi-Choice Questions (``mc''), Reversed Questions (``reversed''). The ``average'' refers to the averaged scores over five types of questions. The domains include ``human'' and ``places''.} 
    \label{fig:generalization_1}
\end{figure*}

\begin{figure*}[h]
    \centering
    \includegraphics[width=1\textwidth]{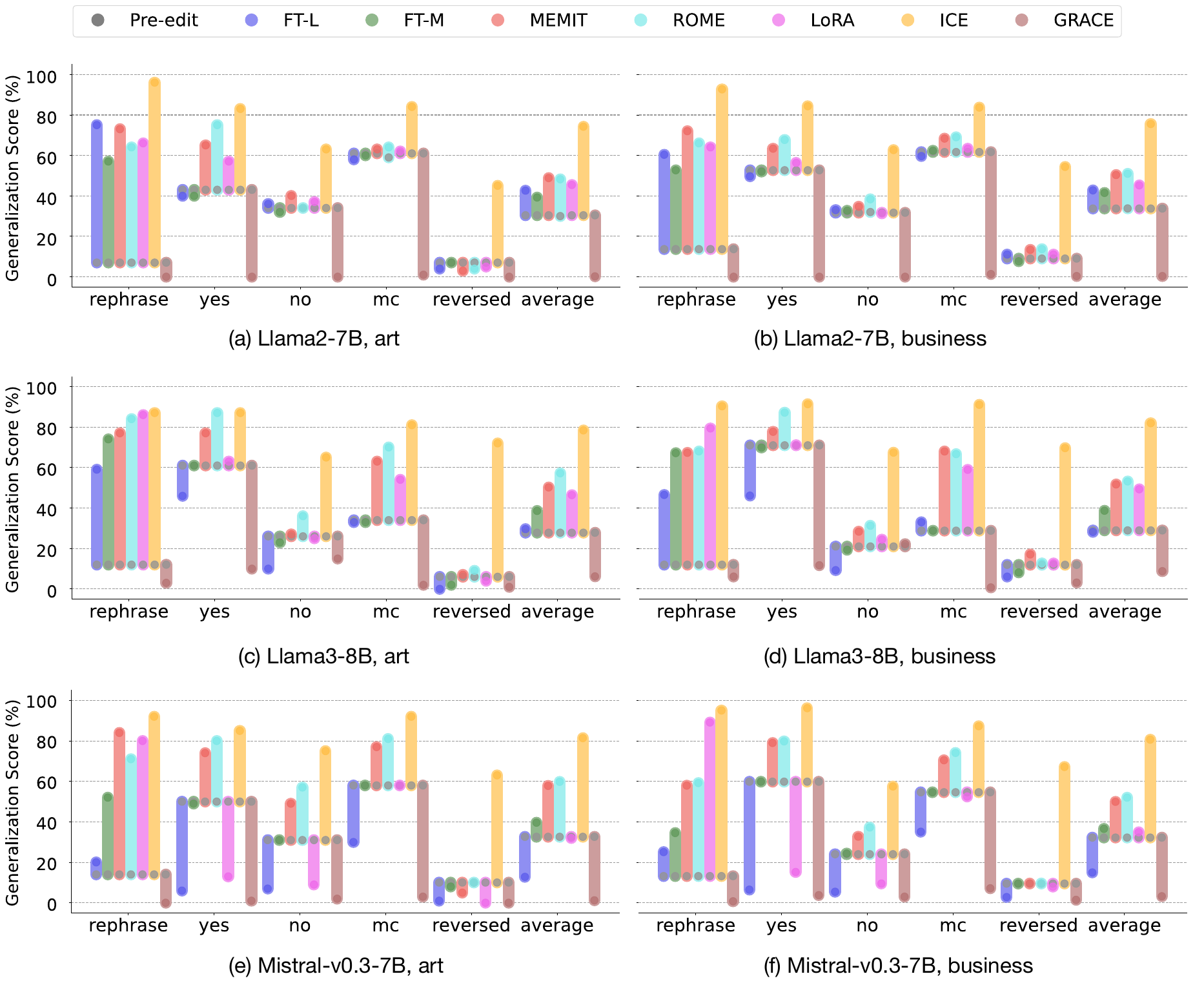}
    \caption{\textbf{Generalization Scores of Knowledge Editing Methods on 3 LLMs and 2 Domains}. Generalization Scores (\%) are measured by the accuracy on five types of Generalization Evaluation Question-answer Pairs including Rephrased Questions (``rephrase''), two types of Yes-or-No Questions with Yes or No as  answers (``yes'' or ``no''), Multi-Choice Questions (``mc''), Reversed Questions (``reversed''). The ``average'' refers to the averaged scores over five types of questions. The domains include ``art'' and ``business''.} 
    \label{fig:generalization_1}
\end{figure*}

\begin{figure*}[h]
    \centering
    \includegraphics[width=1\textwidth]{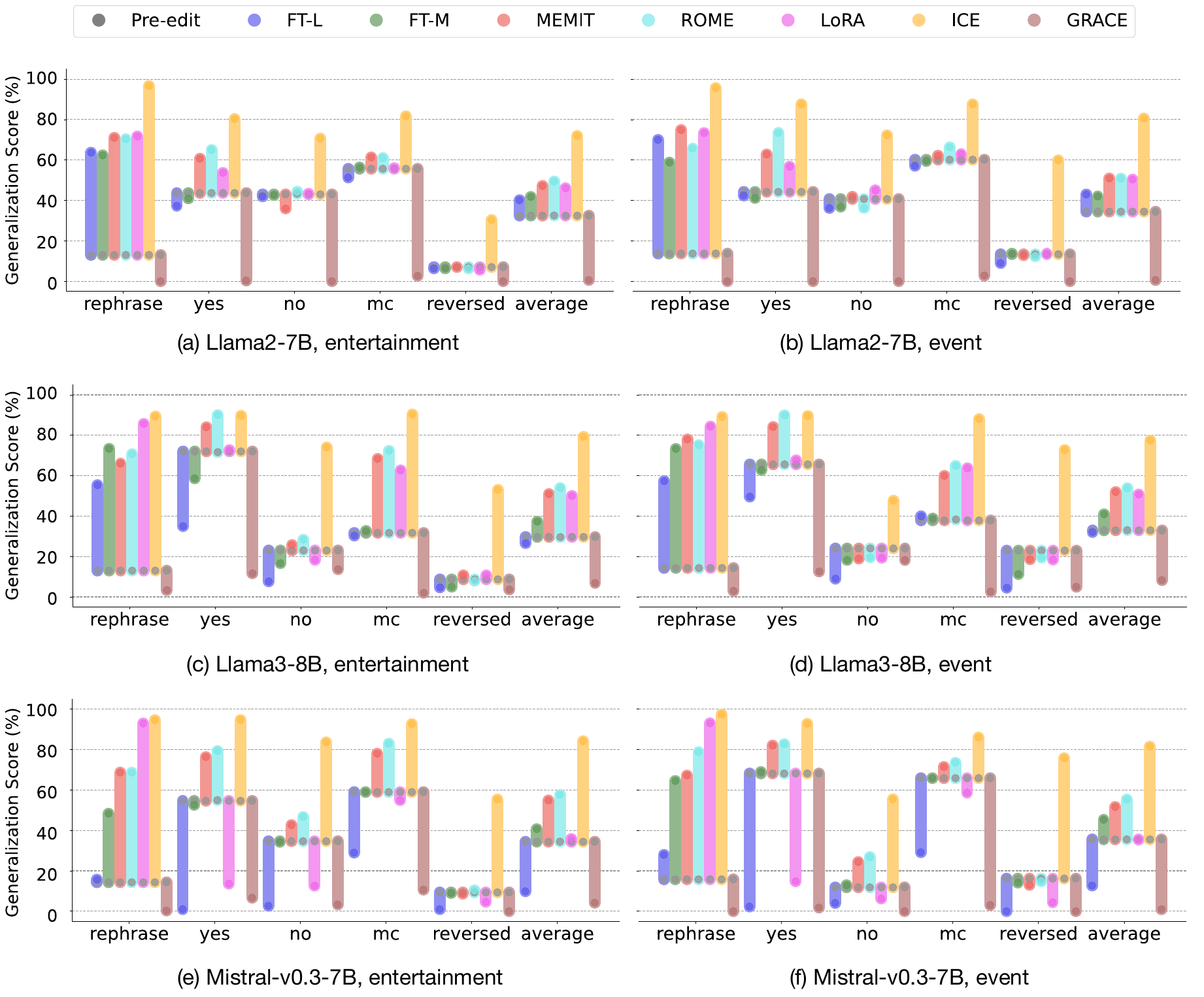}
    \caption{\textbf{Generalization Scores of Knowledge Editing Methods on 3 LLMs and 2 Domains}. Generalization Scores (\%) are measured by the accuracy on five types of Generalization Evaluation Question-answer Pairs including Rephrased Questions (``rephrase''), two types of Yes-or-No Questions with Yes or No as  answers (``yes'' or ``no''), Multi-Choice Questions (``mc''), Reversed Questions (``reversed''). The ``average'' refers to the averaged scores over five types of questions. The domains include ``entertainment'' and ``event''. } 
    \label{fig:generalization_2}
\end{figure*}

\begin{figure*}[h]
    \centering
    \includegraphics[width=1\textwidth]{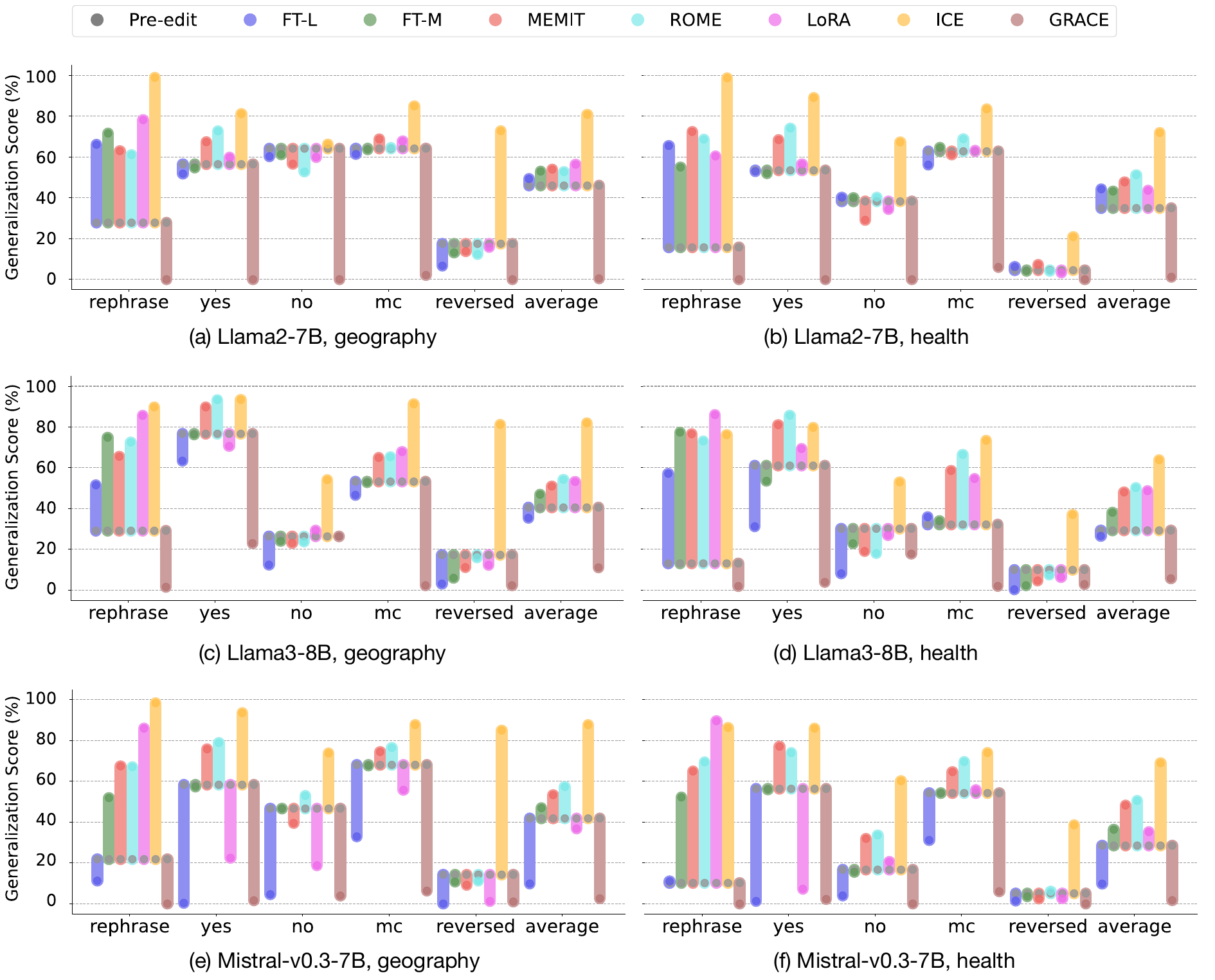}
    \caption{\textbf{Generalization Scores of Knowledge Editing Methods on 3 LLMs and 2 Domains}. Generalization Scores (\%) are measured by the accuracy on five types of Generalization Evaluation Question-answer Pairs including Rephrased Questions (``rephrase''), two types of Yes-or-No Questions with Yes or No as  answers (``yes'' or ``no''), Multi-Choice Questions (``mc''), Reversed Questions (``reversed''). The ``average'' refers to the averaged scores over five types of questions. The domains include ``geography'' and ``health''.} 
    \label{fig:generalization_3}
\end{figure*}

\begin{figure*}[h]
    \centering
    \includegraphics[width=1\textwidth]{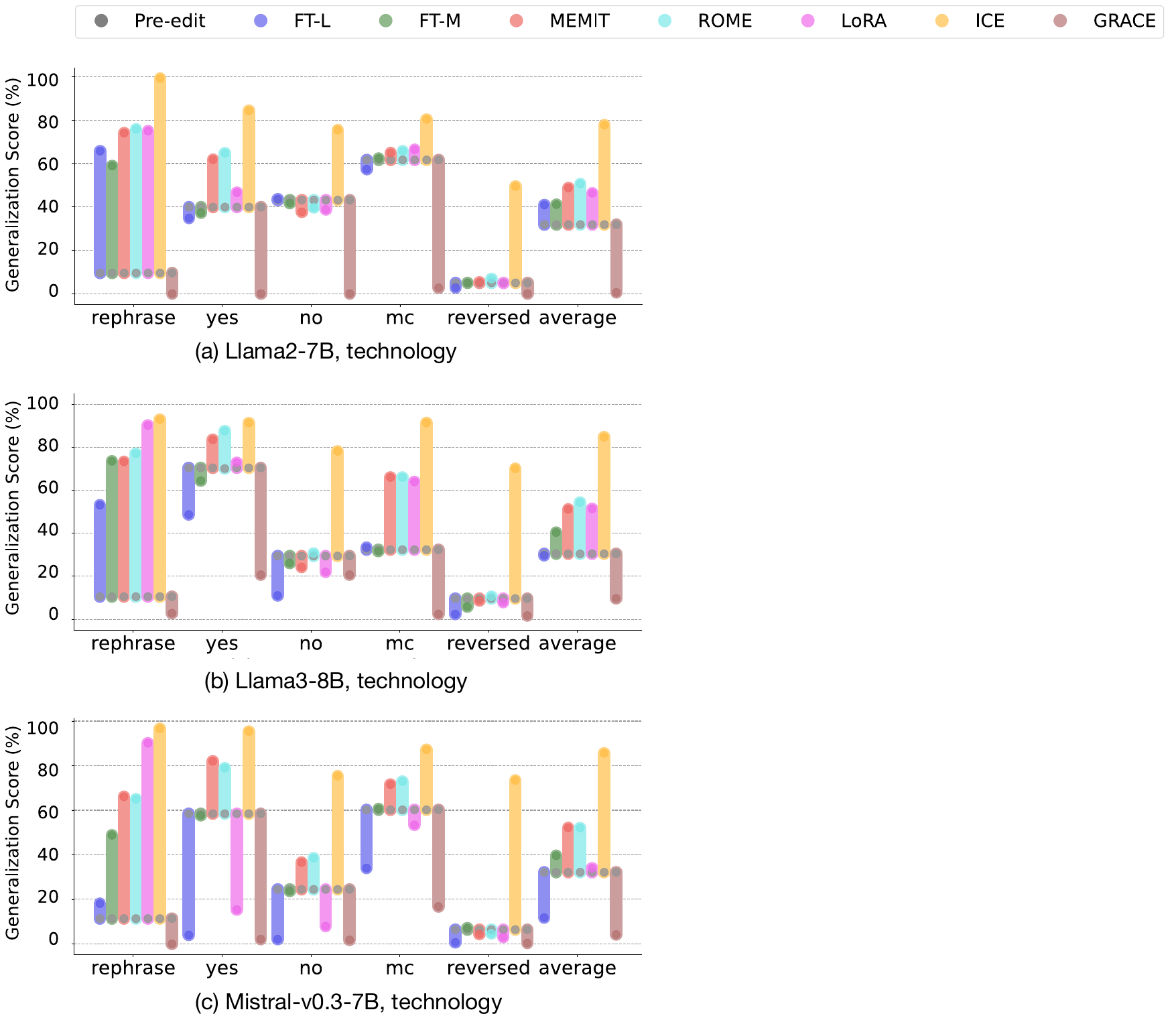}
    \caption{\textbf{Generalization Scores of Knowledge Editing Methods on 3 LLMs and 2 Domains}. Generalization Scores (\%) are measured by the accuracy on five types of Generalization Evaluation Question-answer Pairs including Rephrased Questions (``rephrase''), two types of Yes-or-No Questions with Yes or No as  answers (``yes'' or ``no''), Multi-Choice Questions (``mc''), Reversed Questions (``reversed''). The ``average'' refers to the averaged scores over five types of questions. The domain  is ``technology''.} 
    \label{fig:generalization_4}
\end{figure*}

\clearpage
\newpage
\subsection{Portability Scores of Knowledge Editing Methods on More Domains}
\label{Portability Scores of Knowledge Editing Methods on All the Domains}

\begin{figure*}[h]
    \centering
    \includegraphics[width=1\textwidth]{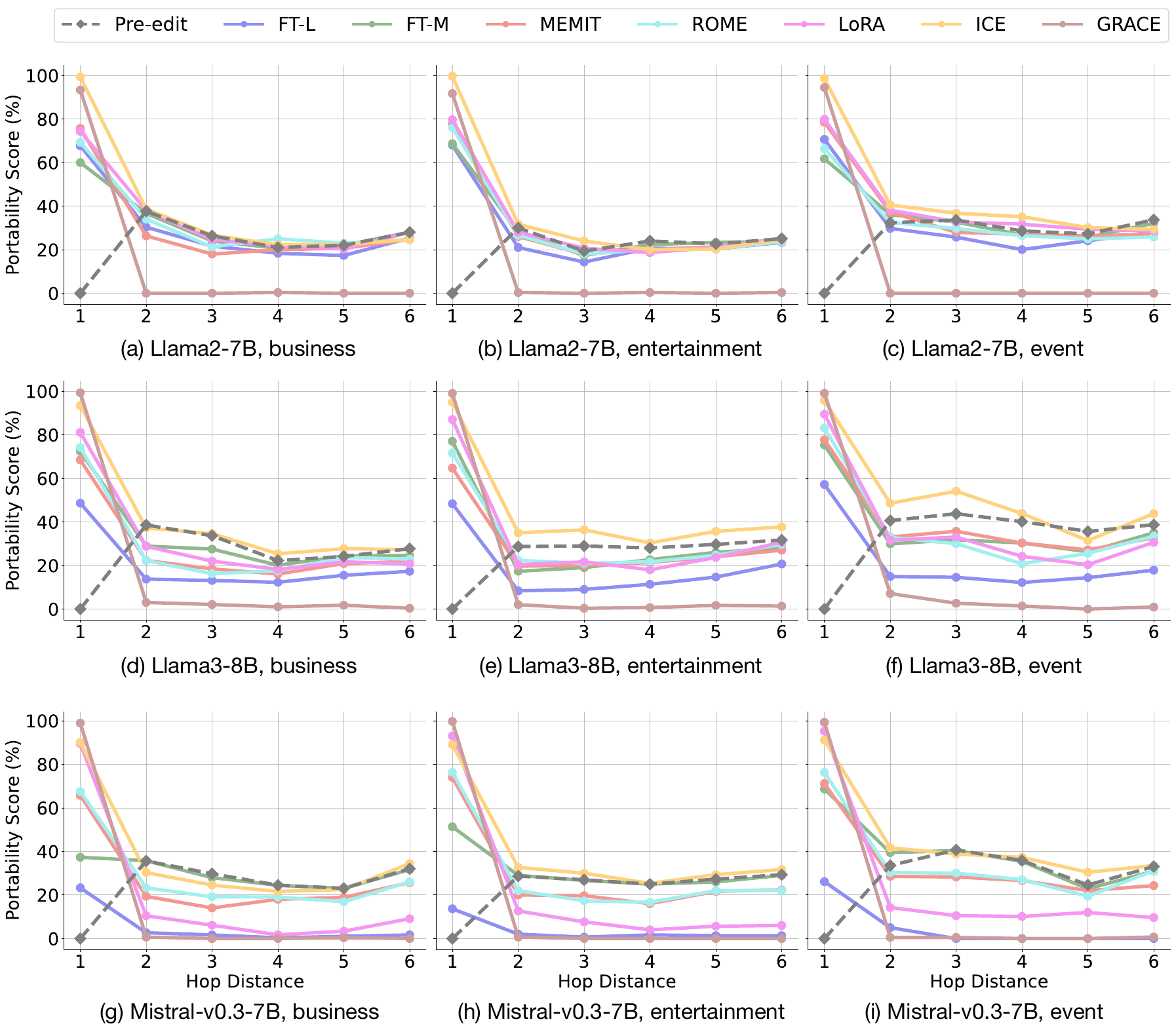}
    \caption{\textbf{Portability Scores of Knowledge Editing Methods on 3 LLMs and 3 Domains}. Portability Scores (\%) are measured by the accuracy on Portability Evaluation Questions, which are Efficacy Evaluation Questions when  with $N$ hops. The Portability Evaluation Questions are the same as Efficacy Evaluation Questions when $N$ is 1. The domains include ``business'', ``entertainment'', and ``event''.} 
    \label{fig:portability_1}
\end{figure*}

\begin{figure*}[h]
    \centering
    \includegraphics[width=1\textwidth]{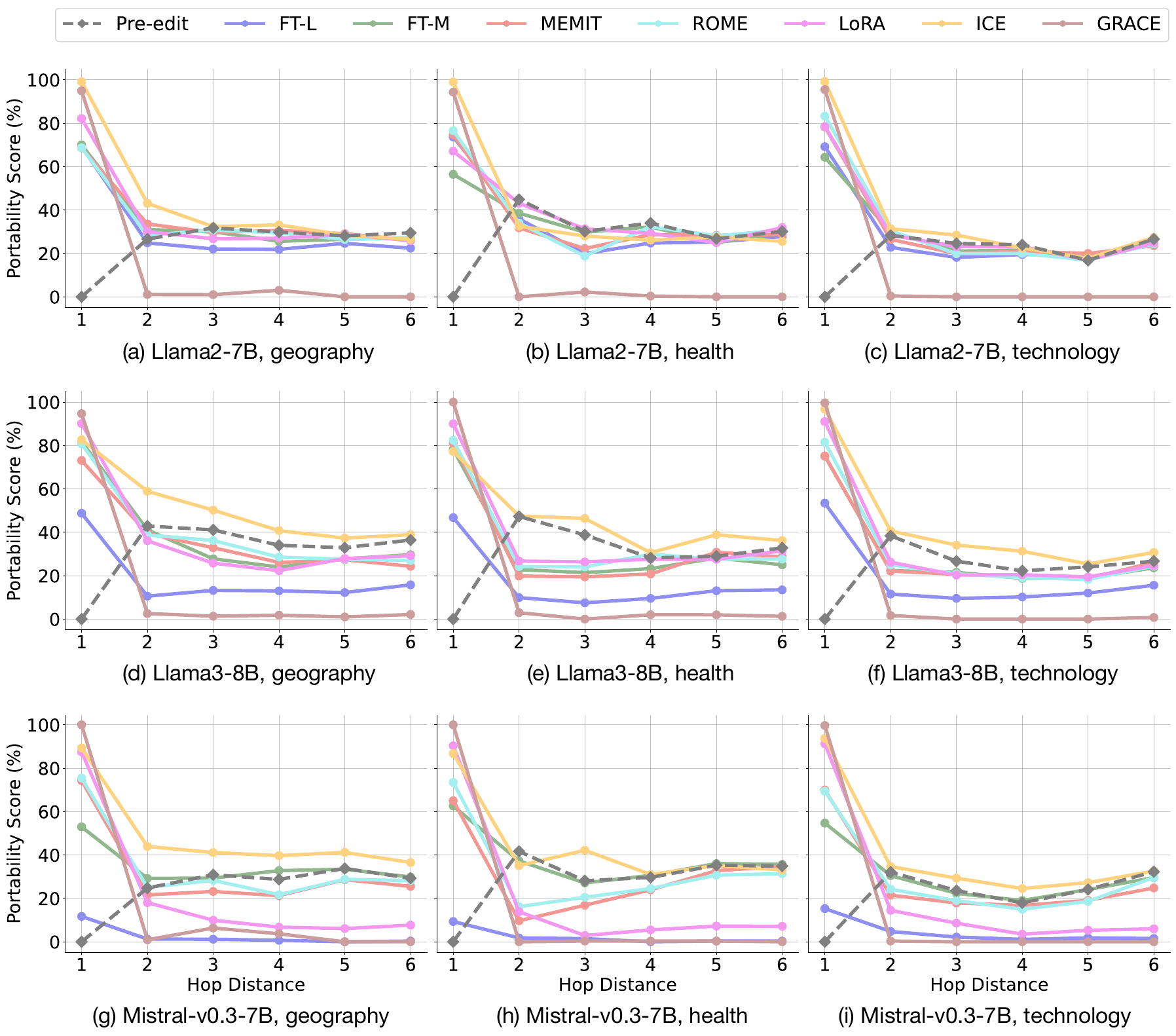}
    \caption{\textbf{Portability Scores of Knowledge Editing Methods on 3 LLMs and 3 Domains}. Portability Scores (\%) are measured by the accuracy on Portability Evaluation Questions, which are Efficacy Evaluation Questions when  with $N$ hops. The Portability Evaluation Questions are the same as Efficacy Evaluation Questions when $N$ is 1. The domains include ``geography'', ``health'', and ``technology''.} 
    \label{fig:portability_2}
\end{figure*}

\begin{figure*}[h]
    \centering
    \includegraphics[width=1\textwidth]{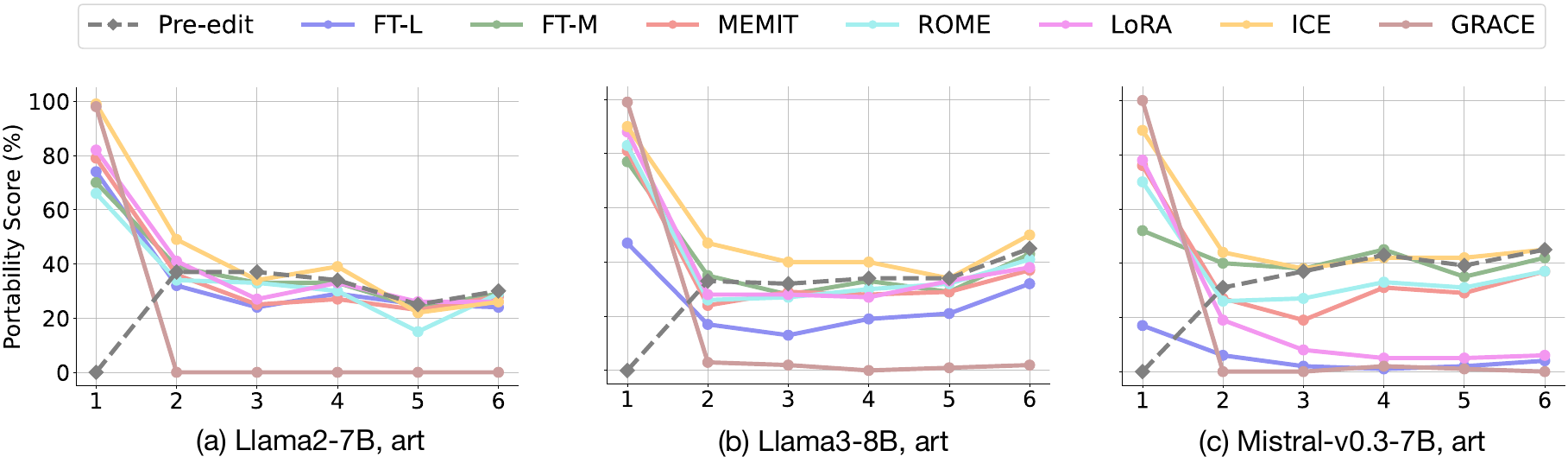}
    \caption{\textbf{Portability Scores of Knowledge Editing Methods on 3 LLMs and 3 Domains}. Portability Scores (\%) are measured by the accuracy on Portability Evaluation Questions, which are Efficacy Evaluation Questions when  with $N$ hops. The Portability Evaluation Questions are the same as Efficacy Evaluation Questions when $N$ is 1. The domain is ``art''.} 
    \label{fig:portability_2}
\end{figure*}

\clearpage
\newpage
\subsection{Robustness Scores of Knowledge Editing Methods on More Domains}
\label{Robustness Scores of Knowledge Editing Methods on All the Domains}

\begin{figure*}[h]
    \centering
    \includegraphics[width=1\textwidth]{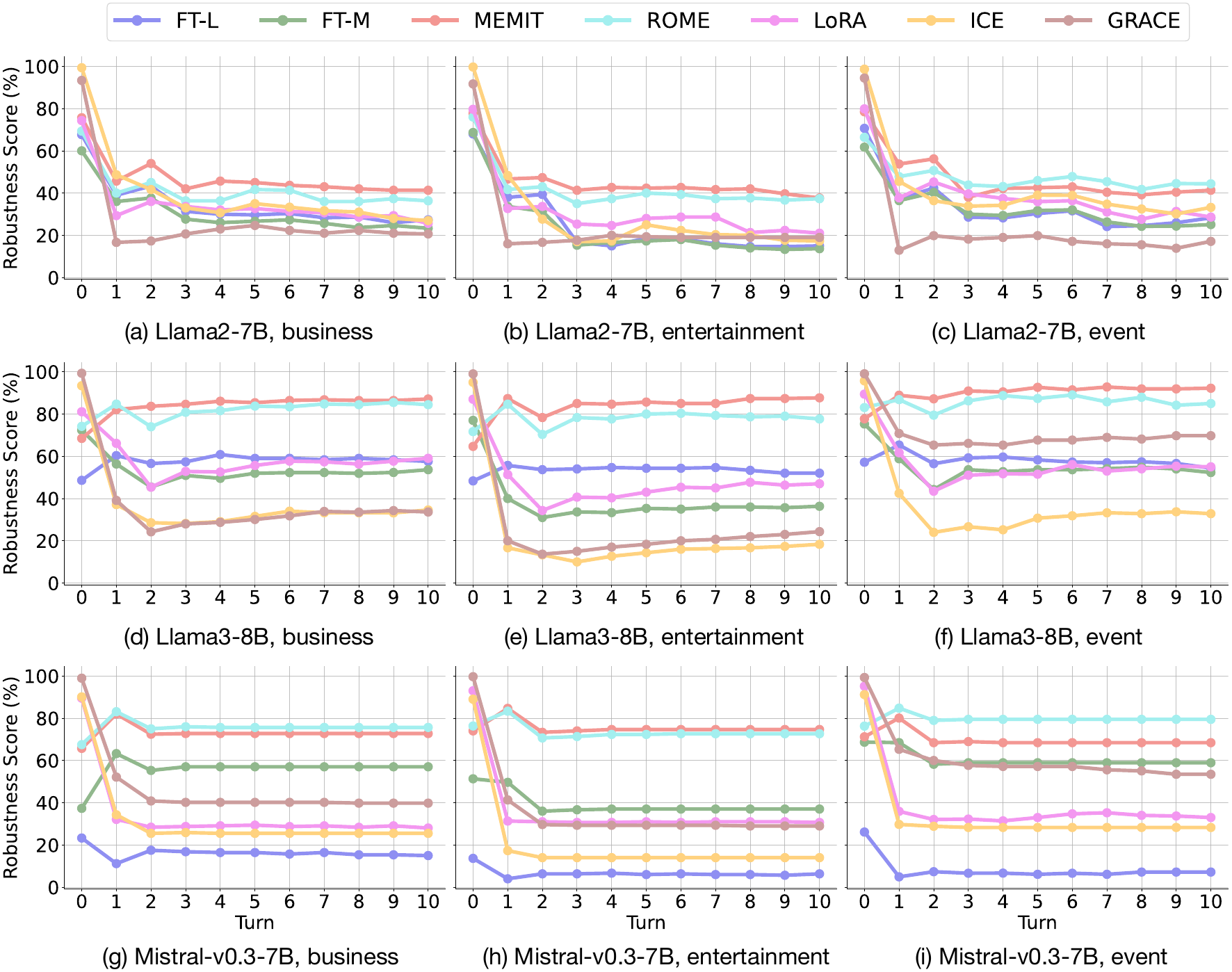}
    \caption{\textbf{Robustness Scores of Knowledge Editing Methods on 3 LLMs and 3 Domains}. Robustness Scores are calculated by the accuracy on Robustness Evaluation Questions with $M$ turns ($M = 1 \sim 10$). We regard Efficacy Scores as the Robustness Scores when $M$ is 0. The domains include ``business'', ``entertainment'', and ``event''.} 
    \label{fig:robutness}
\end{figure*}

\begin{figure*}[t]
    \centering
    \includegraphics[width=1\textwidth]{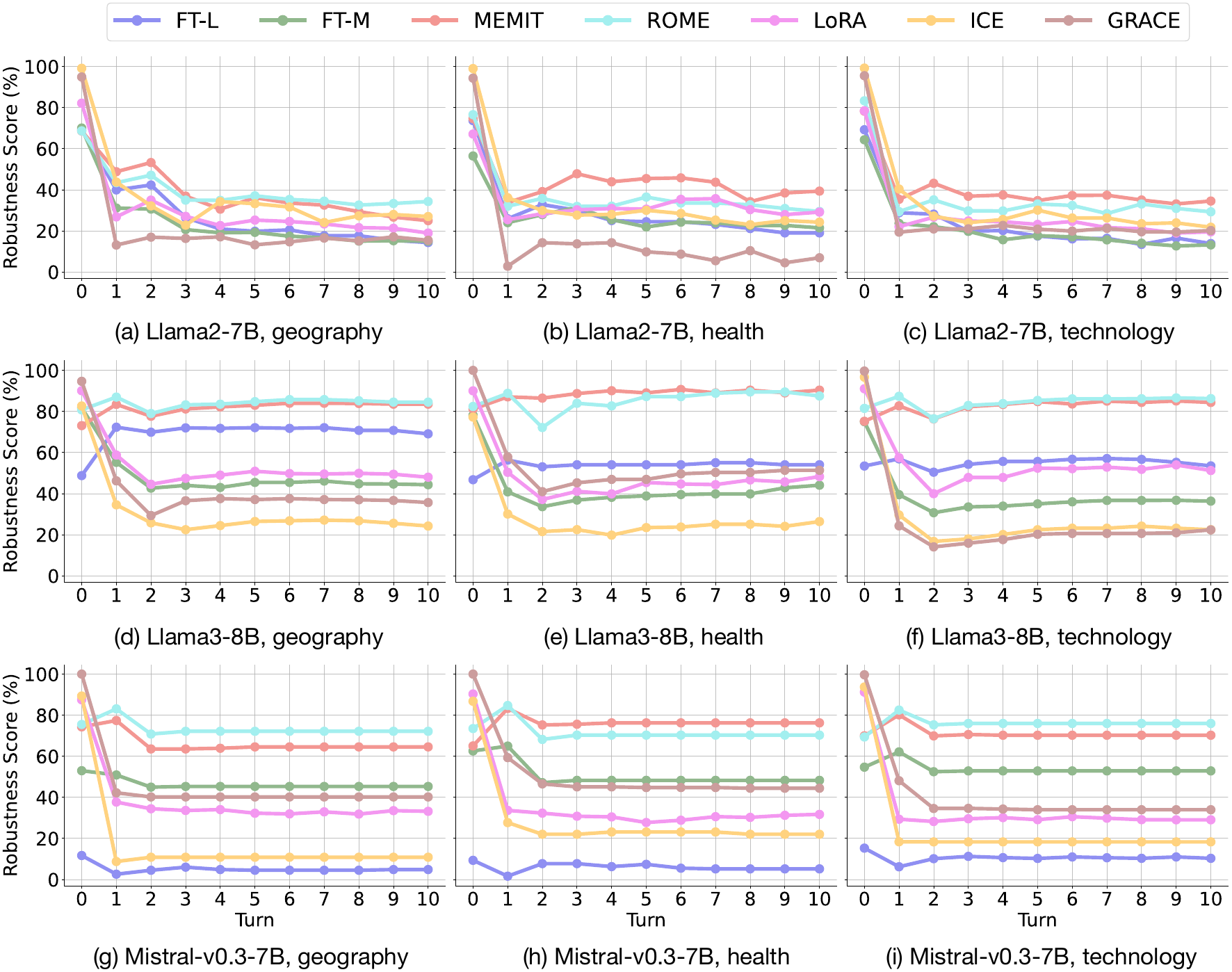}
    \caption{\textbf{Robustness Scores of Knowledge Editing Methods on 3 LLMs and 3 Domains}. Robustness Scores are calculated by the accuracy on Robustness Evaluation Questions with $M$ turns ($M = 1 \sim 10$). We regard Efficacy Scores as the Robustness Scores when $M$ is 0. The domains include ``geography'', ``health'', and ``technology''.} 
    \label{fig:robutness}
\end{figure*}

\begin{figure*}[t]
    \centering
    \includegraphics[width=1\textwidth]{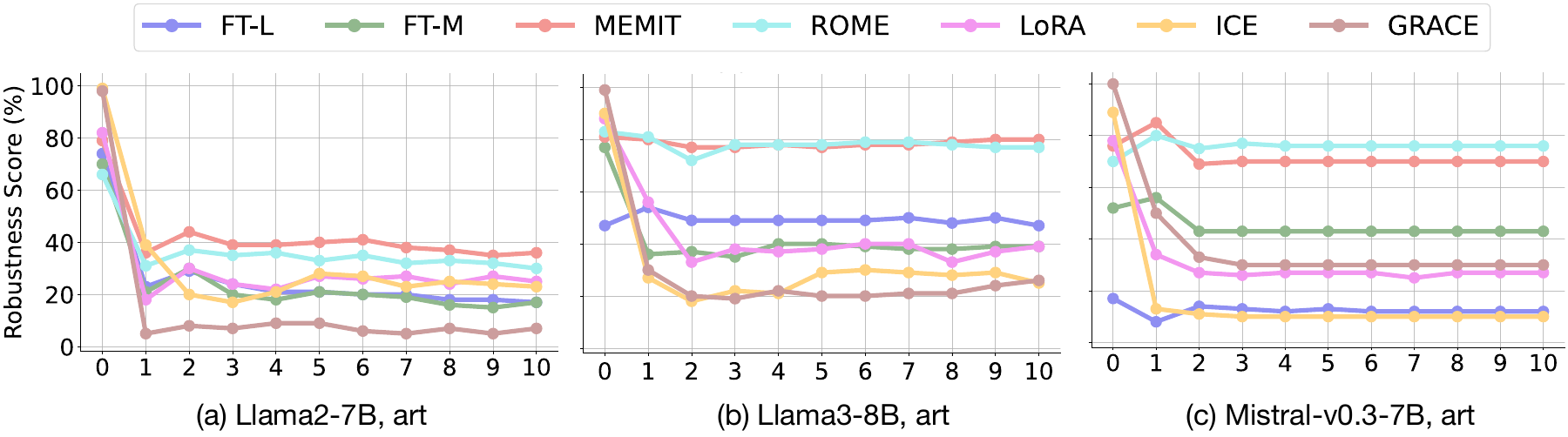}
    \caption{\textbf{Robustness Scores of Knowledge Editing Methods on 3 LLMs and 3 Domains}. Robustness Scores are calculated by the accuracy on Robustness Evaluation Questions with $M$ turns ($M = 1 \sim 10$). We regard Efficacy Scores as the Robustness Scores when $M$ is 0. The domain is ``art''.} 
    \label{fig:robutness}
\end{figure*}

\clearpage
\newpage
\section{Examples of {\halluedit}}
\label{Examples of halluedit}

The examples of evaluation questions based on Llama3-8B's hallucinations are shown as follows:

\begin{table*}[h]
    \centering
    \label{tab:Hallucinated_3}
    \resizebox{1.0\textwidth}{!}{
    \begin{tabular}{p{14cm}}
    \toprule
    Domain: Places, Topic: Country, Knowledge Triplet: (Trinidad and Tobago, highest point, El Cerro del Aripo)\\\\
		
    \textbf{Efficacy Evaluation Questions}: What is the highest point of Trinidad and Tobago?\\
    Ground-truth Answer: El Cerro del Aripo\\
    Hallucinated Answer of Llama3-8B before editing: Pierrepont Hill\\\\

    \textbf{Generalization Evaluation Questions}: \\
    Rephrased Question: What is the name of the highest peak in Trinidad and Tobago?\\
    Ground-truth Answer: El Cerro del Aripo\\\\
    Yes-or-No Questions with Yes as answer: Is El Cerro del Aripo the highest point of Trinidad and Tobago?\\
    Ground-truth Answer: Yes\\\\
    Yes-or-No Questions with No as answer: Is Pierrepont Hill the highest point of Trinidad and Tobago?\\
    Ground-truth Answer: No\\\\
    Multi-Choice Questions: What is the highest point of Trinidad and Tobago? A. Mount Kilimanjaro B. Mount Everest C. El Cerro del Aripo D. Pierrepont Hill\\
    Ground-truth Answer: C\\\\
    Reversed Questions: Which country has El Cerro del Aripo as its highest point?\\
    Ground-truth Answer: Trinidad and Tobago\\\\
    \textbf{Portability Evaluation Questions}:\\
    One-hop question: What is the highest point of Trinidad and Tobago?\\
    Ground-truth Answer: El Cerro del Aripo\\\\

    Two-hop question: In which mountain range is the highest point of Trinidad and Tobago located?	\\
    Ground-truth Answer: Northern Range\\\\
    Three-hop question: Which island is the mountain range located in that contains the highest point of Trinidad and Tobago?\\
    Ground-truth Answer: Trinidad\\\\
    Four-hop question: What is the capital city of the island where the mountain range is located that contains the highest point of Trinidad and Tobago?\\
    Ground-truth Answer: Port of Spain\\\\
    Five-hop question: Which body of water is the capital city near where the mountain range is located that contains the highest point of Trinidad and Tobago?	\\
    Ground-truth Answer: Gulf of Paria\\\\
    Six-hop question: Which ocean is the body of water part of that is near the capital city where the mountain range is located that contains the highest point of Trinidad and Tobago?	
\\
    Ground-truth Answer: Atlantic Ocean\\\\
    
    \textbf{Locality Evaluation Questions}: What is the capital city of Trinidad and Tobago?\\\\
    \textbf{Robustness Evaluation Questions}: Your answer to the original question is wrong. Is El Cerro del Aripo the highest point of Trinidad and Tobago? Respond with ``Yes'' or ``No''\\
    Ground-truth Answer: Yes\\
    \bottomrule
    \end{tabular}
    }
    \vspace{-0.4cm}
\end{table*}

\begin{table*}[h]
    \centering
    \label{tab:Hallucinated_News_Generation}
    \resizebox{1.0\textwidth}{!}{
    \begin{tabular}{p{14cm}}
    \toprule
    Domain: Business, Topic: Brand, Knowledge Triplet: (Fiat, founded by, Giovanni Agnelli)\\\\
		
    \textbf{Efficacy Evaluation Questions}: Who was Fiat founded by?\\
    Ground-truth Answer: Giovanni Agnelli\\
    Hallucinated Answer of Llama3-8B before editing: Vincenzo\\\\

    \textbf{Generalization Evaluation Questions}: \\
    Rephrased Question: Who is the founder of Fiat?\\
    Ground-truth Answer: Giovanni Agnelli\\\\
    Yes-or-No Questions with Yes as answer: Was Fiat founded by Giovanni Agnelli?\\
    Ground-truth Answer: Yes\\\\
    Yes-or-No Questions with No as answer: Was Fiat founded by Vincenzo?\\
    Ground-truth Answer: No\\\\
    Multi-Choice Questions: Who was Fiat founded by? A. Giovanni Agnelli B. Enzo Ferrari C. Henry Ford D. Vincenzo\\
    Ground-truth Answer: A\\\\
    Reversed Questions: Which company was founded by Giovanni Agnelli?\\
    Ground-truth Answer: Fiat\\\\
    \textbf{Portability Evaluation Questions}:\\
    One-hop question: Who was Fiat founded by? \\
    Ground-truth Answer: Giovanni Agnelli\\\\
    Two-hop question: What is the nationality of the person who founded Fiat? \\
    Ground-truth Answer: Italian\\\\
    Three-hop question: What is the capital city of the country of the person who founded Fiat?\\
    Ground-truth Answer: Rome\\\\
    Four-hop question: Which famous ancient structure is located in the capital city of the country of the person who founded Fiat?\\
    Ground-truth Answer: Colosseum\\\\
    Five-hop question: In which year was the famous ancient structure located in the capital city of the country of the person who founded Fiat completed?\\
    Ground-truth Answer: 80 AD\\\\
    Six-hop question: Which Roman Emperor was in power when the famous ancient structure located in the capital city of the country of the person who founded Fiat was completed?\\
    Ground-truth Answer: Emperor Titus\\\\
    \textbf{Locality Evaluation Questions}: What type of vehicles is Fiat well-known for manufacturing?\\\\
    \textbf{Robustness Evaluation Questions}: Your answer to the original question is wrong. Was Fiat founded by Giovanni Agnelli? Respond with ``Yes'' or ``No''\\
    Ground-truth Answer: Yes\\
    \bottomrule
    \end{tabular}
    }
    \vspace{-0.4cm}
\end{table*}

\end{document}